%% file: main.tex
\pdfoutput=1

\documentclass[11pt]{article}

\usepackage[preprint]{acl}

\usepackage{times}
\usepackage{latexsym}
\usepackage{tabularx}
\usepackage{booktabs} 
\usepackage{enumitem} 
\usepackage{tcolorbox}
\tcbuselibrary{listings, breakable}
\usepackage[T1]{fontenc}

\usepackage[utf8]{inputenc}

\usepackage{microtype}

\usepackage{inconsolata}

\usepackage{graphicx}
\usepackage{caption}   
\usepackage{subcaption} 
\usepackage{multirow}
 \usepackage{amsmath} 
\title{Large Language Models with Temporal Reasoning for Longitudinal Clinical Summarization and Prediction} 

\author{
  \textbf{Maya Kruse\textsuperscript{1}},
  \textbf{Shiyue Hu\textsuperscript{1,2}},
  \textbf{Nicholas Derby\textsuperscript{1,2}},
  \textbf{Yifu Wu\textsuperscript{1}},
  \textbf{Samantha Stonbraker\textsuperscript{1}}
\\
  \textbf{Bingsheng Yao\textsuperscript{3}},
  \textbf{Dakuo Wang\textsuperscript{3}},
  \textbf{Elizabeth Goldberg\textsuperscript{1}},
  \textbf{Yanjun Gao\textsuperscript{1}}
\\
\textsuperscript{1}University of Colorado Anschutz Medical Campus \\
  \textsuperscript{2}University of Colorado Boulder \\
  \textsuperscript{3}Northeastern University \\
\\
  \small{
    \textbf{Correspondence:} \href{mailto:yanjun.gao@cuanschutz.edu}{yanjun.gao@cuanschutz.edu}
  }
}

\begin{document}
\maketitle
\begin{abstract}
Recent advances in large language models (LLMs) have shown potential in clinical text summarization, but their ability to handle long patient trajectories with multi-modal data spread across time remains underexplored. This study systematically evaluates several state-of-the-art open-source LLMs, their Retrieval Augmented Generation (RAG) variants and chain-of-thought (CoT) prompting on long-context clinical summarization and prediction. We examine their ability to synthesize structured and unstructured Electronic Health Records (EHR) data while reasoning over temporal coherence, by re-engineering existing tasks, including discharge summarization and diagnosis prediction from two publicly available EHR datasets. Our results indicate that long context windows improve input integration but do not consistently enhance clinical reasoning, and LLMs are still struggling with temporal progression and rare disease prediction. While RAG shows improvements in hallucination in some cases, it does not fully address these limitations. Our work fills the gap in long clinical text summarization, establishing a foundation for evaluating LLMs with multi-modal data and temporal reasoning. 
\end{abstract}

\section{Introduction}

\input{intro}

\section{Related Work}
\input{related}
\section{Dataset and Tasks}
\input{dataset}

\section{Methods}
\input{methods}

\section{Results}

\input{results}

\section{Discussion}

\input{discussion}

\section{Conclusion}

We evaluated LLMs on long-context clinical summarization and prediction using two public EHR datasets. Current models struggle with accurate summarization and temporal reasoning, making it hard to interpret medical event sequences. While RAG offers some improvement, overall performance remains inadequate, highlighting the need for further progress.

\section*{Limitations} 

This study focused on current LLMs' capabilities on the task of clinical summarization for long context documents including temporal information. We evaluated several different models, including Qwen, Llama 3 and DeepSeek, but we acknowledge that this selection is by no means comprehensive and could have limited our analysis. It also does not contain any closed-source LLMs, as the use of these models with MIMIC and EHRShot data is prohibited by Data Use Agreements. Due to the dearth of publicly available EHR datasets, we were only able to include data from two sources: MIMIC-III and EHRShot. 

Our human evaluation was constrained by real-world clinical scheduling limitations, allowing us to consult only one emergency department (ED) physician. Furthermore, we did not formally validate our proposed survey instrument, though the survey questions were aggregated from established prior work (see Appendix regarding the literatures these questions came from). Our goal is to leverage domain experts to better understand LLM limitations, and we plan to expand human evaluation efforts in future studies to provide a more comprehensive assessment.

\section*{Ethical Statement}
This study utilizes de-identified patient data from publicly available datasets (MIMIC-III and EHRShot), ensuring that no identifiable patient information is used. As a result, our work does not involve human subjects research and poses no risk to patient privacy or confidentiality.

Additionally, our study is purely retrospective and computational, focusing on evaluating LLMs for clinical summarization. The models analyzed do not interact with real-world clinical workflows and are not used for actual medical decision-making. Therefore, there is no potential harm to patients or healthcare providers as a result of this research.

Our goal is to assess and improve LLM capabilities for handling complex clinical data, with the long-term aim of developing safe, explainable, and effective AI tools to support healthcare professionals in the future. However, we acknowledge that applying existing LLMs in clinical workflows carries real risks, including privacy leakage, algorithmic biases, and the potential for incorrect decisions. Addressing these challenges is critical for ensuring the safe and ethical deployment of AI in healthcare. 


\bibliography{custom}

\appendix

\section{Appendix}
\label{sec:appendix}

\input{appendix}
\end{document}

%% file: intro.tex
\begin{figure}[ht!]
    \centering
    \includegraphics[width=\columnwidth]{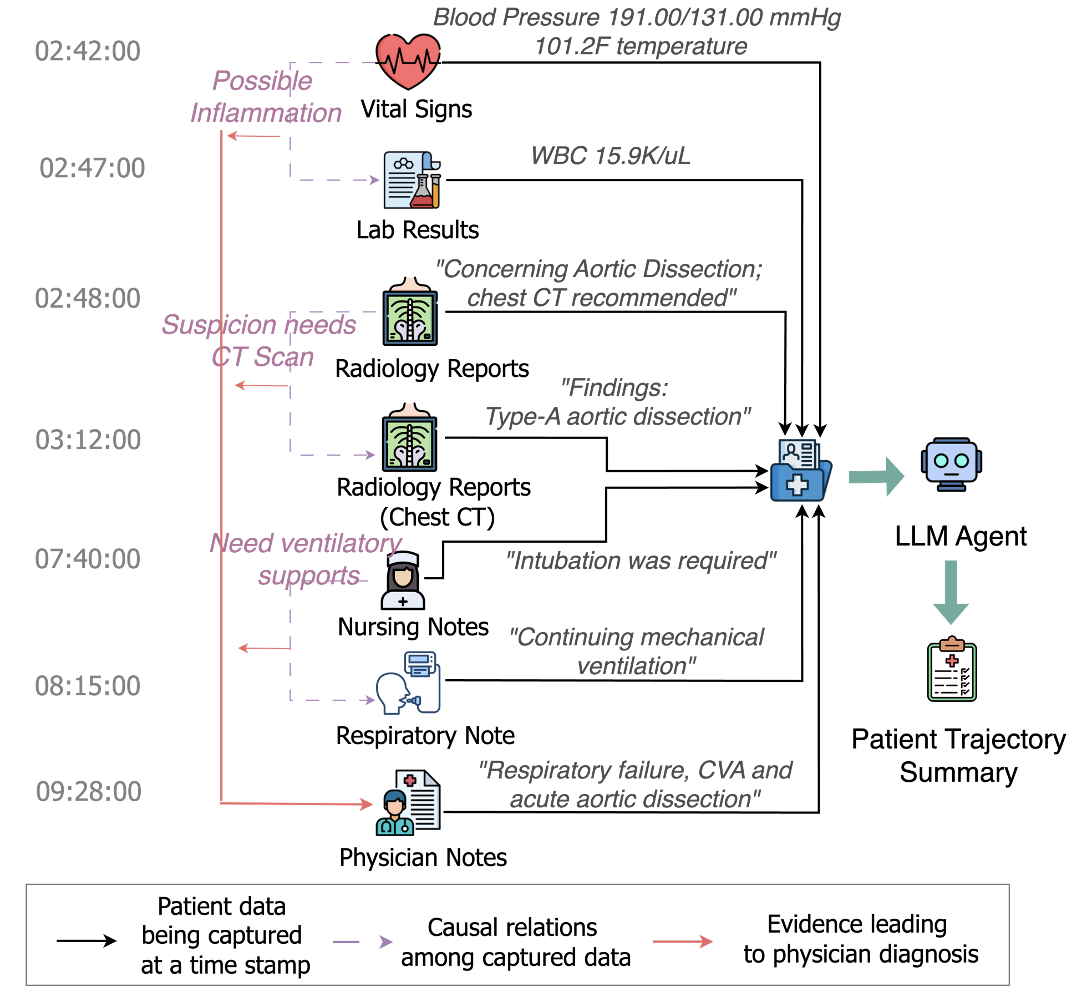}
    \vspace{-.25in}
    \caption{\small An illustration of the longitudinal patient trajectory summarization process from multi-modal EHRs. Key causal relationships among medical observations and interventions are highlighted, leading to the final physician diagnosis. }
    \label{fig:intro}
\end{figure}

Electronic Health Records (EHRs) encapsulate a wide range of multi-modal data, such as vital signs, laboratory results, radiology findings, and free text clinical notes~\cite{mohsen2022artificial,li2022integrating,belden2017dynamic}. Organized across various timestamps, they reflect the dynamic nature of patient care. In clinical settings, particularly for older patients in intensive care units (ICUs) with multiple encounters, chronic conditions, and complex treatment plans, EHRs become especially lengthy and intricate. Clinicians reviewing these long patient histories face significant cognitive burden, often leading to inaccurate decisions such as diagnostic errors~\cite{dymek2021building,singh2017global}. Automated summarization can help improve care continuity and decision-making~\cite{dymek2021building,adams2021s,gao2023overview,laxmisan2012clinical,pivovarov2015automated,liang-etal-2019-novel-system}, but the volume and complexity of EHRs pose challenges, requiring models that handle diverse and evolving clinical data. As shown in Figure~\ref{fig:intro}, summarizing longitudinal patient trajectories requires extracting temporally ordered events from multi-modal data, with temporal reasoning critical for preserving clinical causality. 

Our work evaluates large language models (LLMs) for summarizing complex, longitudinal EHRs, capturing full patient trajectories rather than isolated clinical snapshots. While LLMs have shown promise in clinical NLP tasks~\cite{silcox2024potential,wachter2024will,garcia2024artificial,adams2024speer,gao2022summarizing}, existing evaluations primarily focus on short-context settings and task-specific fine-tuning. Clinical summarization benchmarks, such as discharge summarization (``Discharge Me''\cite{xu-etal-2024-overview}) and diagnosis generation from progress notes (``ProbSum''\cite{gao2023overview}), evaluate LLMs on isolated segments of patient records rather than full hospital trajectories. 
While Retrieval-Augmented Generation (RAG) has been applied to clinical tasks like diagnosis prediction and discharge summarization~\cite{xuretrieval,gao2023retrieval,lewis2020retrieval,myers2024lessons,lyu2024uf}, its effectiveness for long, temporally rich clinical narratives remains unclear.

This work addresses two key gaps: (1) Current evaluations rarely test long-context LLMs in zero-shot settings, leaving their \textbf{inherent capabilities and limitations} in complex medical reasoning unknown; and (2) most clinical benchmarks focus on single time-point summaries, overlooking the demands of \textbf{longitudinal patient care}. We systematically evaluate LLMs on full patient trajectories to assess their ability to process temporally evolving, multi-modal clinical data. 

In this paper, we focus on two public EHR datasets: Medical Information Mart for Intensive Care (MIMIC-III)~\cite{johnson2020mimic} and EHRShot~\cite{wornow2023ehrshot}. We evaluate five state-of-the-art LLMs and their RAG variants: Mistral-7B-Instruct-v0.1 ~\cite{jiang2023mistral}, Llama3-8B-Instruct ~\cite{llama3modelcard}, Qwen2.5-7B ~\cite{yang2024qwen2}, DeepSeek-R1-Distill-Qwen-32B ~\cite{deepseekai2025deepseekr1incentivizingreasoningcapability} and Llama2-13B-chat-hf ~\cite{touvron2023llama}. Our work advances clinical LLM summarization through the following contributions:
\vspace{-0.08in}
\begin{itemize}[leftmargin=*]
\setlength{\itemsep}{-0.3em}
\item We reformulate discharge summarization, progress note summarization, and diagnosis classification into new long-context tasks requiring temporal reasoning, aligned with real clinical workflows.
\item Our tasks integrate structured and unstructured data across multiple timestamps, enabling analysis of \textit{modality} and \textit{temporal context} effects.
\item We compare direct generation, retrieval-augmented generation and chain-of-thought (CoT) prompting approaches for handling long clinical documents.
\end{itemize}
\vspace{-0.04in}

In the absence of benchmarks combining \textit{multi-modal inputs}, \textit{temporal structure}, and \textit{clinical workflow alignment}, our study provides a first step toward evaluating LLMs in real-world longitudinal summarization. The results highlight key limitations and suggest directions for more temporally grounded and clinically usable models. 


%% file: related.tex
\paragraph{Clinical text summarization} Existing work includes discharge summarization~\cite{xu-etal-2024-overview,lyu2024uf}, diagnosis summarization~\cite{gao2023overview,gao2022summarizing,liang-etal-2019-novel-system}, hospital course summarization~\cite{adams2021s,adams2024speer}. While these tasks inherently involve multi-document summarization by human physicians, NLP formulations typically treat them as single-document summarization or multi-document summarization with fixed timestamps (e.g. at the time when patient is discharged). 
\vspace{-.1in}
\paragraph{LLMs for long clinical document} Directly processing the long document input can lead to LLMs' ``lost-in-the-middle" problem~\cite{liu2024lost}. As a result, RAG has been the default when handling long clinical documents, evidenced by its superior performance in diagnosis prediction, discharge summarization and information extraction~\cite{myers2024lessons,lyu2024uf,lopez2025clinical}. Due to the task setup, these works have only investigated single modality EHRs with long input length. 

Our work addresses this problem by extending RAG and CoT-based approaches to multi-modal EHR summarization and clinical prediction, incorporating both structured (tabular) and unstructured (clinical notes) data to enhance the completeness and accuracy of patient trajectory summaries. We evaluate how well LLMs and their RAG set ups can capture and preserve temporal and causal relationships across patient's hospital stay, bridging the gap between isolated document processing and comprehensive longitudinal understanding.  


 

%% file: dataset.tex

We use two complementary datasets, MIMIC-III and EHRShot, to evaluate LLMs' clinical reasoning. 
They differ in three key aspects:
 1) Modalities: MIMIC-III includes both structured data and clinical notes, while EHRShot contains only structured data.
 2) Output type: MIMIC-III tasks focus on generating short summaries of patient progress or discharge status, whereas EHRShot involves classification-based diagnosis prediction.
 3) Decision time span: MIMIC-III targets immediate ICU-related summaries, while EHRShot predicts diagnoses within a year of discharge, emphasizing long-term forecasting based on prior visits.
 Despite these differences, both datasets address longitudinal patient information, requiring models to reason over time and synthesize complex clinical histories.

\subsection{MIMIC-III}
MIMIC-III is a publicly available dataset comprising de-identified health records from over 40,000 ICU patients. For this study, we focus on a subset of patients with hospital stays exceeding 72 hours to ensure sufficient context for evaluating long-document summarization and temporal reasoning. Unlike prior MIMIC-III studies, our selected cohort emphasizes complex, multi-day ICU stays where accurate summaries are most impactful. This setup allows us to assess LLMs’ ability to handle prolonged and evolving clinical narratives.

We use both structured and unstructured data. Specifically, we extract chart events (vital signs, ventilator settings), lab events (e.g., white blood cell counts), input events (e.g., IV infusions, feedings), and medications. These features reflect key aspects of clinical decision-making and support rich, temporally grounded summarization tasks.

\paragraph{Discharge Summarization}
Discharge summaries are a crucial part of a patient’s hospital care process, providing a comprehensive overview of their hospital stay and key clinical events. It is usually written by the physician after the patient is discharged, and contains three main sections: \textsc{Diagnosis}: a list of diagnoses requiring an understanding of the patient's clinical progression up to discharge; \textsc{Brief Hospital Course}: clinical summary of treatments, interventions, and significant events during hospitalization; \textsc{Discharge Instructions}: post-discharge guidance, including medication plans, dietary recommendations and follow-up care. 




Discharge Summarization has been explored in several studies, including ~\cite{xu2024overview} and ~\cite{ando2022artificial}, often focusing on specific sections like the Brief Hospital Course or discharge instructions. In this paper, we generate \textit{all three sections} by extracting and chronologically ordering structured and unstructured data from a hospital admission. We limit input to the last 24 hours to prevent overloading the LLM while prioritizing the most relevant information for discharge, considering the generally long hospital stays ($\geq$3 days). We also test a 48-hour window to capture a broader clinical context.  

We design four input settings to evaluate LLMs on multi-modal and temporal reasoning: \textsc{note only} where input only contains clinical notes, \textsc{tabular only} where input only contains tabular data, \textsc{shuffled tabular} where we shuffled the tabular data by their timestamps, \textsc{combined} where we combined both notes and tabular data, sorted by their timestamps. 

\paragraph{Assessment and Plan (A\&P) Generation}
\begin{figure}[htp]
    \centering
    \begin{subfigure}{\columnwidth}
    \small 
        \centering
        \resizebox{\columnwidth}{!}{ 
        \begin{tabular}{@{}p{1.5cm}p{5.5cm}@{}}
        \toprule
        \textbf{Symbol} & \textbf{Definition} \\ \midrule 
        \( D_i \) & Progress note for day \( i \) \\ 
        \( X_i \) & Input data used to generate the progress note for day \( i \) \\ 
        \midrule
        \textbf{Method} & \textbf{Input Data for Day \( i \)} \\ 
        \midrule
        \textbf{Baseline} & \( X_i = \text{EHR}_i \) (No prior progress notes) \\ 
        \textbf{Single-Day Context} & \( X_i = (\text{EHR}_i, D_{i-1}) \) (Includes previous day's note) \\ 
        \textbf{Multi-Day Context} & \( X_i = (\text{EHR}_i, D_1, D_2, ..., D_{i-1}) \) (Includes all previous notes) \\ 
        \bottomrule
        \end{tabular}
        }
    \end{subfigure}
    \vspace{.05em}
        \begin{subfigure}{\columnwidth}
        \centering
\includegraphics[width=\columnwidth]{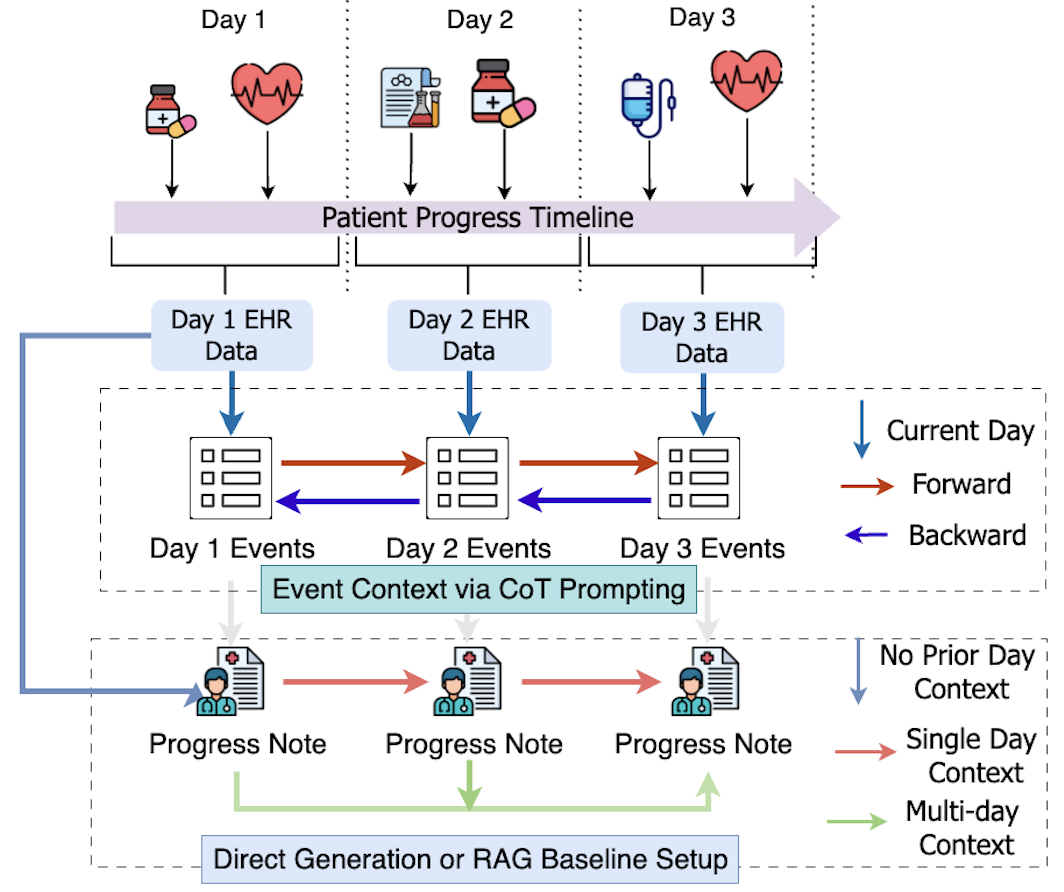}
    \end{subfigure}
    
    \vspace{-.12in}
    \caption{\small Illustration of the A\&P generation workflow (top) and a comparison of input data formulations (bottom).}
    \label{fig:combined_pn}
\end{figure}


Daily progress notes are the documents where physicians record diagnoses, treatment, and clinical status, providing key insights into a patient’s condition throughout their hospital stay. A progress note typically consists of four sections: Subjective, Objective, Assessment and Plan~\cite{weed1968medical,wright2014bringing}. 
The Subjective and Objective sections serve as the "evidence" and observation of the patient on that day, with Subjective comprising unstructured free text describing symptoms, status, and treatment, while Objective consists of structured tabular data like lab results and charted values. In contrast, the Assessment and Plan sections capture the physician's reasoning and clinical hypothesis based on this evidence, with diagnoses and treatment plans listed. To assess LLM abilities in summarizing longitudinal data, the Assessment and Plan sections provide a great testbed, requiring integration of information from multiple days, capturing the evolution of the patient's condition, and synthesizing key clinical evidence into a concise yet comprehensive summary.

\begin{figure}[t]
    \includegraphics[width=\columnwidth]{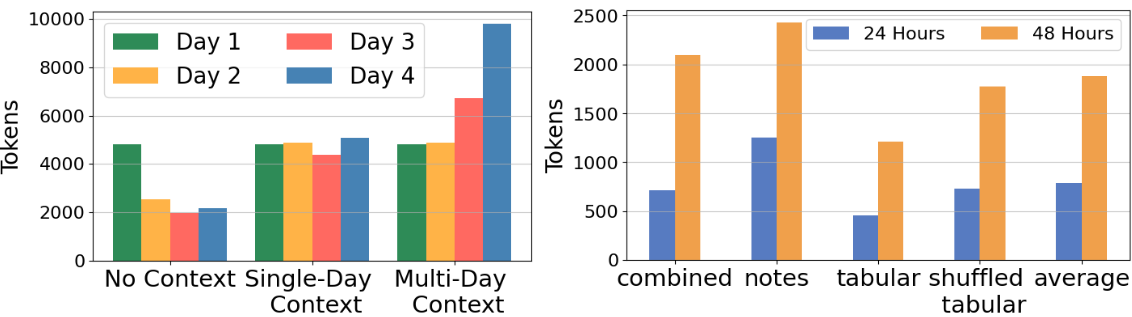}
    \vspace{-.25in}
    \caption{\small The token length of input data used for A\&P generation across days and methods (left), as well as that of input data used for Discharge Summarization across time windows and modalities(right).}
    \label{fig:tokens}
\end{figure}

The setup of the task is illustrated in Figure~\ref{fig:combined_pn}, which explains the three input methods: the \textsc{no prior context} (\textsc{Baseline}) method uses only the current day's structured EHR data, the \textsc{Single-Day Context} method adds the previous day's progress note for limited historical context, and the \textsc{Multi-Day Context} method incorporates all prior progress notes for a more comprehensive patient history. 
For day $i$, the input includes progress notes up to $i-1$ (depending on the method), but the note for $i$ is excluded and compared to the generated version. This setup reflects the real clinical workflow, where physicians reference prior notes when writing new ones. 



\vspace{-0.5em}
\paragraph{How long is the input in MIMIC?}  Figure~\ref{fig:tokens} shows input lengths across tasks and settings. For discharge summarization, inputs average 500 tokens (24-hour window) and 1,940 tokens (48-hour window). In contrast, the Assessment and Plan (A\&P) generation task involves much longer inputs due to the inclusion of prior progress notes and structured data. The No Context baseline starts with a relatively high token count (3,125 tokens on Day 1) but remains the shortest overall. Single-Day Context, which adds the previous day’s note, increases input length to 5,375 tokens. Multi-Day Context, which accumulates all prior notes, shows the steepest token growth—reaching nearly 10,000 tokens by Day 4 and averaging 6,875 tokens.




\subsection{Diagnosis Prediction from EHRShot}
EHRShot~\cite{wornow2023ehrshot} is a longitudinal dataset comprising fully structured EHR data from 6,739 patients at Stanford Medicine, with over 41 million clinical events. In contrast to MIMIC-III, which contains a mix of structured and unstructured data, EHRShot offers a clean, structured-only setting, allowing for controlled evaluation of temporal reasoning in longitudinal patient records.

We include EHRShot to complement the MIMIC-based summarization tasks in two important ways. First, EHRShot focuses on future prediction—each task requires predicting whether a patient will receive one of six diagnoses (e.g., \textsc{Hypertension}, \textsc{Pancreatic Cancer}, \textsc{Acute MI}) within one year after discharge. This differs from our MIMIC tasks, which emphasize retrospective summarization of observed hospital stays. Second, its purely structured nature allows us to isolate models’ abilities to handle temporal patterns in tabular data without the added complexity of clinical language. Together, these contrasts offer a broader assessment of LLM capabilities in modeling temporal EHR information across different modalities and predictive targets.

The six diagnosis tasks vary in prevalence from common conditions like \textsc{Hyperlipidemia} (31.85\%) to rarer ones like \textsc{Celiac} (3.46\%). Inputs have a mean token length of 1,989 {\footnotesize($\sigma$: 1,270)}, with a median of 1,851, and an average of 74 unique measurements per patient. More statistics are provided in Appendix~\ref{sec:ehrshot_stat}. 


\begin{table}[ht!]
\small	
\centering
\begin{tabular}{@{}p{7.5cm}@{}} 
\toprule
\textbf{First entry:} Urea Nitrogen is 26 mg/dL. Hematocrit is 34.20\%. Hemoglobin is 12.20 g/dL. 12 minutes later: Glucose is 214 mg/dL. Lactate is 2.40 mmol/L. \textbf{3 minutes later:} Hemoglobin is 13.00 g/dL. Lactate is 2.40 mmol/L. pO2 is 441 mm Hg. \textbf{1 hour later}: 100.00 ml of 0.9\% Normal Saline is administered. Neosynephrine-k is administered. \textbf{49 minutes later:} Radiology note: 12:48 PM CHEST (PORTABLE AP) Clip Reason: line placement, r/o PTx Admitting Diagnosis: HEAD BLEED; MEDICAL CONDITION: 68-year-old man s/p MVA significant head trauma, intubated s/p r subclavian triple lumen placement... \\
\bottomrule 
\end{tabular}
\vspace{-.1in}
\caption{\small An example of a patient's compiled data, including relative timestamp (minutes and hours since previous recorded data), structured data converted into narrative format as well as unstructured note data.}
\label{tab:chronology}
\end{table}

\subsection{Representing multi-modal, longitudinal patient data }
We convert structured EHR data into natural language to enable LLM-based summarization, following prior table-to-text approaches~\cite{gao2024raw,yu2023unified}. As in Table~\ref{tab:chronology}, each measurement is verbalized using simple templates (e.g., {\footnotesize \textsc{[Measurement] is [Value][Unit]}}), grouped by timestamp, and temporally ordered using relative time references. Medications and input events follow a similar format (e.g., “is administered”).

To reduce redundancy from repeated or copy-pasted entries, we deduplicate records across modalities and retain only the most recent entry when multiple identical values appear within a short time window. Clinical notes are filtered similarly. For EHRShot, structured inputs are divided into six hour chunks and repetitive phrases are removed. Appendix~\ref{sec:more_preprocessing}. 

%% file: methods.tex
We evaluate five LLMs with varying context capacities. LLaMA2-13B and Mistral-7B support up to 4K tokens, while LLaMA3-8B allows for 8K. To test long-context capabilities, we include Qwen2.5-7B and DeepSeek-R1 (32B), both of which support input lengths up to 128K tokens. While we did explore biomedical LLMs (PMC-LLaMa, BioGPT, BioMistral), their performance was extremely poor, so we did not include them in this paper.




\subsection{Approaches}
We compare a baseline direct generation, structured RAG and CoT event extraction approach. The key difference between these approaches lies in how they handle long temporal contexts. Direct generation processes the entire input at once but may suffer from the lost-in-the-middle problem, where relevant information buried in long documents is overlooked. In contrast, RAG segments temporal information into retrievable chunks, which helps manage long sequences but may disrupt temporal dependencies between events. For CoT event extraction, the long context is first compressed into a series of clinical events, over which the model then summarizes.
\vspace{-0.5em} \paragraph{1). Direct Generation.}
The patient’s chronological data, formatted according to the specific task requirements, is provided to the model as a direct input without any additional structural modifications. This serves as a baseline to assess the capability of current state-of-the-art language models in processing and generating clinical summaries from raw sequential data, without the aid of task-specific adaptations or architectural enhancements. 
\vspace{-0.5em} 
\paragraph{2). RAG.}
 For this setup, we choose the same selection as models used for direct generation, with the exception of DeepSeek, as it is considerably larger than the other LLMs (32B parameters vs 7-13B) yet fails to outperform these smaller models. This combination of high computational cost and suboptimal performance led us to exclude DeepSeek for RAG. \citeauthor{myers2024lessons} (\citeyear{myers2024lessons}) studied the quality of embedding in medical RAG and found BGE~\cite{bge_embedding} yielded the highest performance. Thus, we adopt BGE embedding and apply it to all LLM RAG set ups in our work. Additionally, the queries presented in the paper are slightly modified and used to perform query optimization. Hyperparameter optimization is carried out on the standard RAG hyperparameters chunk size, chunk overlap and top-k retrieved documents. 

\vspace{-0.5em}
\paragraph{3). Event Extraction via Chain-of-Thought (CoT).} As an alternative to retrieval or direct generation, we introduce a chain-of-thought (CoT)~\cite{wei2022chain} prompting approach that incorporates a clinical reasoning step before summarization. LLMs are prompted to first identify temporally ordered clinical events from the input data, which are appended to the original input as structured context for generation. This intermediate step emphasizes salient clinical signals and reduces input noise, helping the model focus on meaningful patterns. 

We develop our prompt with guidance from recent studies showing that LLMs perform better on clinical tasks when given clear, structured instructions tailored to the specific goal. For example, \citet{wang2023art} finds that prompts targeting specific event types like symptoms, lab results, treatments, and medical decisions help models focus on clinically meaningful information and improve event detection. \citet{yuan-etal-2023-zero} also shows that when complex tasks are broken into smaller reasoning steps, especially for understanding how events are ordered over time, models tend to produce more consistent and accurate outputs. Figure~\ref{fig:cot_prompt} illustrates the CoT prompt design for A\&P generation task.

\begin{figure}[t]
\centering
\small 
\begin{tcolorbox}[title=\textbf{Chain-of-Thought Prompt for ICU Daily Event Extraction}, colback=gray!5, colframe=gray!40!black, width=\columnwidth]
\textbf{ICU DAILY EVENT EXTRACTION TASK}

Analyze this structured ICU data by identifying critical clinical events, paying special attention to numerical values and their progression. Only report values showing meaningful change or clinical significance. For repeated values, mention only those demonstrating changes.

\textcolor{red}{\{Input Text\}}

Identify (with direct references to data points when possible):
\begin{enumerate}[nosep, leftmargin=*] 
    \item Major symptoms or changes (new, worsening, improving) – Specify relevant numerical changes.
    \item Critical test results (labs, imaging, etc.) – Highlight significant abnormal or normal values.
    \item Important treatments or interventions – Clearly link to the preceding clinical data.
    \item Significant care team decisions – Support with relevant clinical data.
    \item Major medical decisions or diagnoses – Reference pertinent clinical observations.
\end{enumerate}

\textbf{Response Format:}

\#\#\# Day X Key Events \#\#\#

- [Time] | [Event Description] (Explanation for identifying this event)
\end{tcolorbox}

\vspace{-.1in}

\begin{tcolorbox}[title=\textbf{Example Model Output}, colback=blue!2, colframe=blue!50!black, width=\columnwidth]
\#\#\# Day 3 Key Events \#\#\#

- \textbf{2178-02-11 09:50:00} | Blood pressure readings of 88/46 mmHg (This low blood pressure reading is concerning and may require additional fluid resuscitation or adjustment of vasoactive medications.)
\end{tcolorbox}
\vspace{-.1in}
\caption{\small Chain-of-Thought prompting template and example output used for ICU daily event extraction from structured EHR data for A\&P generation.}
\label{fig:cot_prompt}
\end{figure}

We experiment with seven temporal input configurations for event extraction: using only the current day’s data, combining it with previous days (\textsc{Forward}), or with subsequent days (\textsc{Backward}), each spanning 1, 3, or all available days. We apply these settings to a development batch of 20 patients and found that, within each respective group of experiments, the \textsc{Forward} setting using all previous days and the \textsc{Backward} setting using one following day produced the best results. Based on these findings, we report performance under three representative temporal contexts: \textsc{+0} (Current Day Only, baseline), \textsc{+n} (Forward Context with all prior days), and \textsc{–1} (Backward Context with one following day). 

\vspace{-0.5em}
\paragraph{Experiment settings.} For all LLMs, we run their 8-bit quantized version. We set the output token length as 1,000, but almost all tasks output is significantly shorter than this limit. For RAG setup, we used Langchain~\cite{langchain} with Faiss~\cite{douze2024faiss} for semantic retrieval in a vector database. We perform hyperparameter tuning with chunk size between [250, 750], top k between [10, 50], chunk overlap between [50,200]. The detailed results of hyperparameter searching are in Appendix~\ref{sec:hyperparam_tuning}. All experiments are run on 2 H100 94GB GPUs. 

\subsection{Evaluation}
On MIMIC, we report standard summarizatoin evaluation metrics that capture string overlap and semantics similarity. ROUGE-L ~\cite{lin2004rouge} measures string overlap, while BERTScore \cite{zhang2019bertscore} assesses maximum token pairwise similarity. We use SapBERT as the backend for BERTScore due to its superior performance in biomedical entity representation~\cite{liu2020self}. On EHRShot, given that the diagnosis prediction tasks are essentially binary classification, we report macro-average accuracy and F-scores. 

%% file: results.tex
The results are organized by task: Discharge Summarization and Assessment and Plan Generation performed on MIMIC-III and Diagnosis Prediction on EHRShot. Additionally, prompt optimization was carried out on all tasks, using a small sample set (Appendix~\ref{sec:prompt_opt}).

\begin{table}[]
\centering
\resizebox{\columnwidth}{!}{  
\begin{tabular}{@{}llll@{}}
\toprule
\textbf{Setting} & \textbf{Model} & \textbf{ROUGE-L} & \textbf{BERTScore} \\
\midrule
\multirow{5}{*}{\begin{tabular}[c]{@{}l@{}}Direct \\ Gen\end{tabular}} 
& Mistral        & 16.28 {\scriptsize ±12.83} & 65.23 {\scriptsize ±13.38} \\
& Llama3         & 11.82 {\scriptsize ±14.42} & 55.35 {\scriptsize ±22.48} \\
& Qwen           & 15.28 {\scriptsize ±5.38}  & 63.98 {\scriptsize ±13.50} \\
& DeepSeek       & 13.51 {\scriptsize ±5.04}  & 63.64 {\scriptsize ±13.86} \\
& Llama2         & 15.34 {\scriptsize ±5.90}  & 63.82 {\scriptsize ±11.44} \\
\midrule
\multirow{4}{*}{RAG}                                                          
& Mistral        & 15.04 {\scriptsize ±2.34}  & 65.89 {\scriptsize ±5.69} \\
& Llama3         & 15.90 {\scriptsize ±2.00}  & 64.38 {\scriptsize ±4.81} \\
& Qwen           & \textbf{17.91} {\scriptsize ±2.35}  & 65.25 {\scriptsize ±4.77} \\
& Llama2         & 16.98 {\scriptsize ±1.59}  & \textbf{67.20} {\scriptsize ±6.21}  \\
\bottomrule
\end{tabular}
}
\vspace{-.1in}
\caption{\small Results on discharge summarization, comparing the direct generation and RAG approaches across all models.}
\label{tab:dg-rag}
\end{table}

\begin{table}[t]
\small
\centering
\resizebox{\columnwidth}{!}{
\begin{tabular}{@{}llcc@{}}
\toprule
\textbf{Section} & \textbf{Method} & \textbf{ROUGE-L} & \textbf{BERTScore} \\
\midrule
\multirow{2}{*}{Dx} 
    & Direct Gen &  \textbf{3.42} {\scriptsize ± 4.34} & \textbf{50.07} {\scriptsize ± 10.78} \\
    & CoT Prompt  & 2.95 {\scriptsize ± 3.48} & 48.46 {\scriptsize ± 11.00} \\
\midrule
\multirow{2}{*}{HC} 
    & Direct Gen  & \textbf{12.28} {\scriptsize ± 3.13} & \textbf{62.51} {\scriptsize ± 5.95} \\
    & CoT Prompt & 9.98 {\scriptsize ± 3.84}  & 59.60 {\scriptsize ± 7.32} \\
\midrule
\multirow{2}{*}{DI} 
    & Direct Gen & \textbf{12.07} {\scriptsize ± 3.90} & 60.05 {\scriptsize ± 9.09} \\
    & CoT Prompt & 10.83 {\scriptsize ± 4.58} & \textbf{60.52} {\scriptsize ± 7.86} \\
\bottomrule
\end{tabular}
}
\vspace{-0.1in}
\caption{\small Mistral performance on the three sections of discharge summarization (Dx: Diagnosis, HC: Hospital course, DI: Discharge Instruction), comparing direct generation approach with the event extraction CoT. }
\label{tab:mistral_direct_vs_cot}
\end{table}


\begin{table*}[ht]
\centering
\label{tab:results}
\resizebox{\textwidth}{!}{%
\begin{tabular}{@{}lcccccccccccc@{}}
\toprule
\multirow{3}{*}{\textbf{Setting}} 
& \multicolumn{6}{c}{\textbf{\textit{Direct Generation}}} 
& \multicolumn{6}{c}{\textbf{\textit{RAG}}} \\
\cmidrule(lr){2-7} \cmidrule(lr){8-13}
& \multicolumn{2}{c}{Mistral} 
& \multicolumn{2}{c}{Llama3} 
& \multicolumn{2}{c}{Qwen} 
& \multicolumn{2}{c}{Mistral} 
& \multicolumn{2}{c}{Llama3} 
& \multicolumn{2}{c}{Qwen} \\
\cmidrule(lr){2-13}
& RL & BS & RL & BS & RL & BS & RL & BS & RL & BS & RL & BS \\
\midrule
No Prior    
& \textbf{17.70}\scriptsize$\pm$3.79 & 68.16\scriptsize$\pm$6.91
& 16.60\scriptsize$\pm$3.39 & 72.01\scriptsize$\pm$6.04 
& 15.15\scriptsize$\pm$3.53 & 72.17\scriptsize$\pm$4.51 
& 17.15\scriptsize$\pm$5.60 & 68.57\scriptsize$\pm$6.16
& 15.48\scriptsize$\pm$3.60 & \textbf{73.78}\scriptsize$\pm$4.50
& {15.60}\scriptsize$\pm$3.60 & 72.16\scriptsize$\pm$4.63 \\
Single-Day  
& 26.11\scriptsize$\pm$11.68 & 71.63\scriptsize$\pm$10.86 
& \textbf{33.16}\scriptsize$\pm$10.68 & 80.10\scriptsize$\pm$8.93
& 20.74\scriptsize$\pm$4.19 & 78.01\scriptsize$\pm$4.95 
& 29.50\scriptsize$\pm$12.46 & 74.23\scriptsize$\pm$7.79 
& 22.09\scriptsize$\pm$5.51 & 78.80\scriptsize$\pm$4.96 
& 25.42\scriptsize$\pm$8.08 & \textbf{79.27}\scriptsize$\pm$5.37 \\
Multi-Day   
& 23.32\scriptsize$\pm$13.05 & 72.41\scriptsize$\pm$9.47 
& \textbf{32.42}\scriptsize$\pm$12.14 & \textbf{80.42}\scriptsize$\pm$8.81 
& 21.60\scriptsize$\pm$5.23 & 78.41\scriptsize$\pm$4.81 
& 27.40\scriptsize$\pm$10.08 & 73.55\scriptsize$\pm$8.46 
& 21.42\scriptsize$\pm$6.81 & 78.96\scriptsize$\pm$5.42 
& 25.36\scriptsize$\pm$7.50 & {80.28}\scriptsize$\pm$4.68 \\
\bottomrule
\end{tabular}%
}
\vspace{-.1in}
\caption{\small Performance of Direct Generation and RAG methods on A\&P generation across models and input settings. RL = ROUGE-L score; BS = BERTScore computed using SapBERT embeddings.} 
\label{tab:ap-transposed}
\end{table*}

\begin{table}[t]
  \centering
  \resizebox{\columnwidth}{!}{%
    \begin{tabular}{l l c c c}
      \toprule
      \textbf{Model} & \textbf{Metric} & \textsc{Current Day} & \textsc{Forward+N} & \textsc{Backward-1} \\
      \midrule
      \multirow{2}{*}{Mistral} 
        & ROUGE-L & 19.85\scriptsize$\pm$10.87 & 20.82\scriptsize$\pm$12.02 & 20.10\scriptsize$\pm$11.15 \\
        & BERTScore & 68.63\scriptsize$\pm$11.56 & 69.69\scriptsize$\pm$10.82 & 69.35\scriptsize$\pm$9.67 \\
      \midrule
      \multirow{2}{*}{Llama3} 
        & ROUGE-L & \textbf{31.56}\scriptsize$\pm$16.56 & 31.34\scriptsize$\pm$16.93 & \textbf{28.40}\scriptsize$\pm$14.78 \\
        & BERTScore & \textbf{77.89}\scriptsize$\pm$12.48 & 76.17\scriptsize$\pm$14.74 & 74.66\scriptsize$\pm$16.31 \\
      \midrule
      \multirow{2}{*}{Qwen}   
        & ROUGE-L & 20.41\scriptsize$\pm$4.39  & 20.69\scriptsize$\pm$4.65 & 20.70\scriptsize$\pm$4.70 \\
        & BERTScore & 77.44\scriptsize$\pm$4.17  & 77.61\scriptsize$\pm$4.27 & \textbf{77.10}\scriptsize$\pm$4.40 \\
      \bottomrule
    \end{tabular}%
  }
  \vspace{-.1in}
  \caption{\small Event extraction CoT results on A\&P generation task across settings.}
  \label{tab:event-extraction-transposed}
\end{table}

\begin{figure}[ht!]
    \centering
    \includegraphics[scale=0.38]{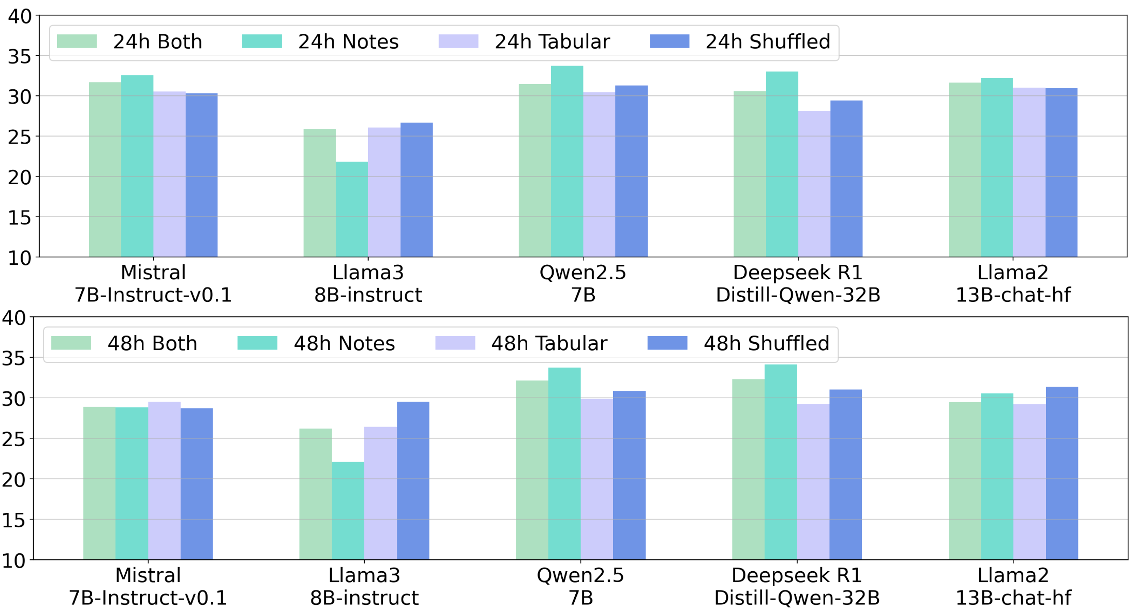}
    \vspace{-.25in}
    \caption{\small Average f1 results across modalities and time windows (on direct generation)} 
    \label{fig:average_f1}
\end{figure}

\begin{table}[]
\centering
\resizebox{\columnwidth}{!}{%
\begin{tabular}{llllll}
\toprule
\textbf{}                          & \textbf{} & \multicolumn{2}{c}{\textbf{Direct Gen}} & \multicolumn{2}{c}{\textbf{RAG}} \\
\textbf{}                          & \textbf{} & \textbf{Mistral}     & \textbf{Qwen}    & \textbf{Mistral} & \textbf{Qwen} \\ \hline
\multirow{2}{*}{Acute MI}          & Accuracy  & 52.84                & 62.72            & 63.03            & 65.50         \\
                                   & F1        & 33.45                & 2.58             & 11.83            & 16.87         \\ \hline
\multirow{2}{*}{Celiac Disease}    & Accuracy  & 62.96                & 96.05            & 95.04            & 95.30         \\
                                   & F1        & 2.60                 & 0.00             & 0.00             & 9.52          \\ \hline
\multirow{2}{*}{Hyperlipidemia}    & Accuracy  & 58.02                & 68.15            & 66.09            & 63.03         \\
                                   & F1        & 30.33                & 1.53             & 12.74            & 7.45          \\ \hline
\multirow{2}{*}{Hypertension}      & Accuracy  & 64.44                & 69.38            & 68.81            & 68.81         \\
                                   & F1        & 26.53                & 6.06             & 8.7              & 8.7           \\ \hline
\multirow{2}{*}{Lupus}             & Accuracy  & 66.42                & 95.06            & 94.79            & 94.81         \\
                                   & F1        & 10.53                & 0.00             & 0.00             & 0.00          \\ \hline
\multirow{2}{*}{Pancreatic Cancer} & Accuracy  & 73.83                & 78.52            & 78.86            & 79.05         \\
                                   & F1        & 13.11                & 2.25             & 0.00             & 0.00          \\ \bottomrule
\end{tabular}%
}
\vspace{-.1in}
\caption{\small Results on the binary diagnosis prediction task using EHRShot data (macro-average).}
\label{tab:ehrshot}
\end{table}

\begin{figure*}[t!]
    \centering
    \includegraphics[scale=0.40]{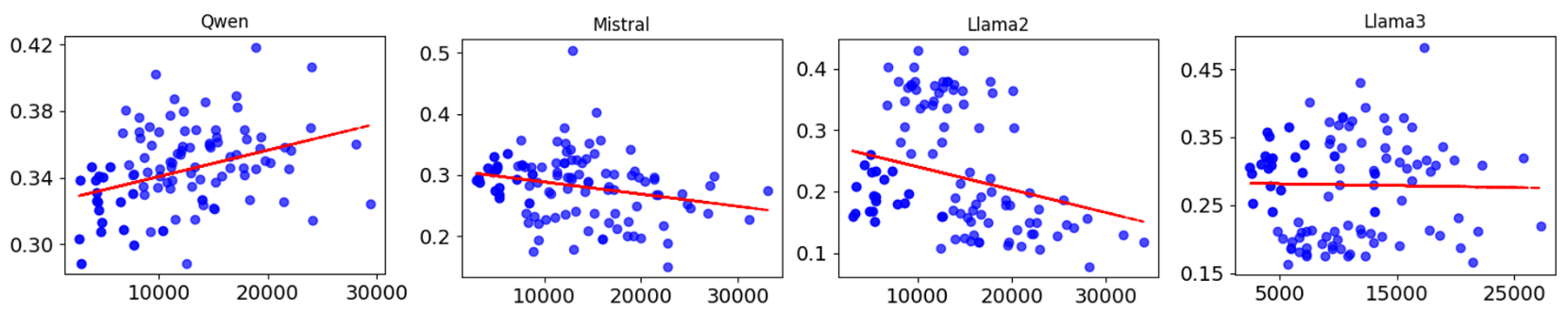}
    \vspace{-.1in}
    \caption{\small Correlation of token length and average performance across all tasks and metrics} 
    \label{fig:scatter}
\end{figure*} 

\begin{figure}[t!]
    \centering
    \includegraphics[width=\columnwidth]{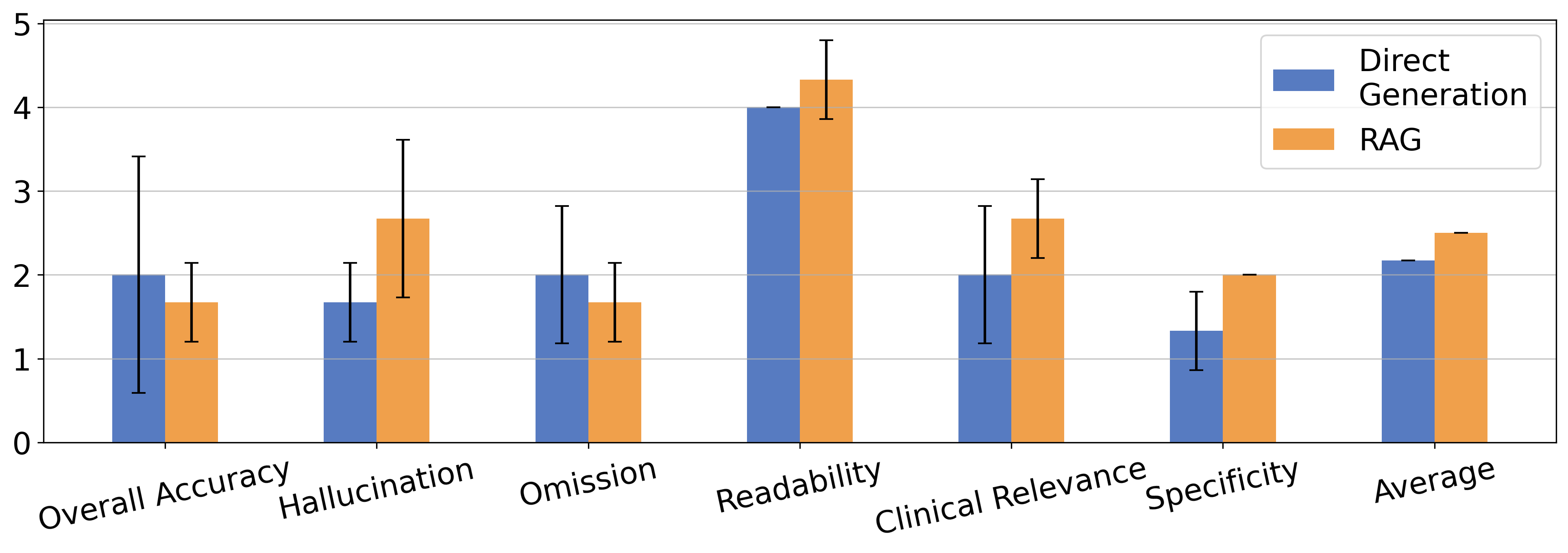}
    \vspace{-.28in}
    \caption{\small Expert review scores for Qwen direct generation and RAG on Discharge Summarization} 
    \label{fig:human-eval}
\end{figure}

\noindent \textbf{Discharge Summarization.}
Table \ref{tab:dg-rag} presents results from comparison between the Direct Generation and RAG approaches. Qwen achieves the highest ROUGE-L score of 17.91 using RAG. In general, most LLMs record a performance increase in their RAG variants, the biggest of which is Llama3’s, with an increase of 4.08 on ROUGE-L. Mistral is the exception, with a minor performance decrease of -1.24. DeepSeek, as the largest LLM, reports mediocre performance

Table~\ref{tab:mistral_direct_vs_cot} shows section-wise analysis for Mistral comparing direct generation and CoT methods, where CoT event extraction performs slightly worse than direct generation across most sections. Other LLMs exhibit similar trends, so we report Mistral only for brevity. In Appendix~\ref{tab:ds_sections}, we provide more section-wise analysis on the Discharge Summ task. Results further confirm the challenge of LLMs reasoning over patient trajectories. 

Figure \ref{fig:average_f1} shows the effects of the four modalities (notes + tabular data, notes only, tabular only and shuffled tabular) and two context window sizes (24 and 48 hours) on performance. For the 24 hour window, the pure note modality dominates, performing best on Mistral, Qwen, DeepSeek and Llama2. Llama3 is the outlier and reports the best performance on the combined and shuffled tabular modalities. In the 48h window, trends are less clear, but most models still favor notes.  

Temporal order appears unhelpful, as shuffled tabular data performs slightly better than chronological tabular data, likely because the data is near the discharge state, where patients’ clinical conditions stabilize, leading to fewer changes over time. 


\noindent \textbf{A\& P Generation.}
Table~\ref{tab:ap-transposed} shows that adding prior context (Single- or Multi-Day) improves performance across all models, particularly for ROUGE scores. Llama3 performs best on direct generation, especially in Single- and Multi-Day Contexts, but is outperformed by Qwen and Mistral on RAG. However, Multi-Day Context does not always yield performance gains over Single-Day, suggesting models struggle with longer patient histories. RAG generally improves performance, particularly for Mistral and Qwen, but its impact varies across models: Llama3 performs better without retrieval. 

This task specifically requires clinical reasoning over past captured data (evidence) to summarize patient progression and plan for treatment, yet the relatively low ROUGE and BERTScore values indicate that models still struggle with temporal reasoning and integrating historical context effectively. 

Table~\ref{tab:event-extraction-transposed} presents results of the event extraction CoT method. CoT did not yield improvements over direct generation or RAG approaches across ROUGE-L or BERTScore. This suggests that, while the event extraction step offers interpretability and control, it may not directly enhance generation quality when applied in a pipeline without further model adaptation. We view this as a useful diagnostic approach that could be further refined or integrated with instruction tuning in future work.

\noindent \textbf{Diagnosis Prediction.}
We evaluate Mistral and Qwen for this task based on their prior performance. Both models prioritize majority class predictions, leading to inflated accuracy but severely low F1-scores. RAG improves accuracy but does not meaningfully enhance recall or F1-score.

The task remains highly challenging, especially for rare diseases, where models struggle to predict positive cases. Qwen achieves high accuracy but an F-score of zero for celiac disease and lupus, reflecting severe class imbalance. It mostly predicts negatives, occasionally misclassifying positives but failing to identify true cases. Given that 96.54\% of celiac and 95.56\% of lupus cases are true negatives, its accuracy (96.05\% and 95.06\%) is barely lower than predicting all negatives. 



\paragraph{Overall Performance} Figure~\ref{fig:scatter} illustrate how input length impacts model performance across different LLMs, considering their varying context windows. Qwen (128K context), shows a slight positive correlation, suggesting that longer inputs may improve its performance. In contrast, Mistral and Llama2 (both 4K context), decline as input length increases, likely due to exceeding their optimal processing capacity. Llama3 (8K context), remains stable with minimum impact from the length. Models with shorter contexts struggle with longer inputs, while larger-context models handle them better, though benefits remain inconsistent.  



%% file: discussion.tex

Another critical aspect to consider is running time and memory usage. All 7B LLMs require approximately 15GB of GPU RAM for direct generation. RAG demands significantly more memory, peaking at 32–33GB, nearly twice that of direct generation. For Qwen, it takes 18-20 minutes to run RAG on one progress note with prior context input, while direct generation takes 10 minutes on average. On EHRShot where input is shorter, diagnosis prediction takes just 3–4 seconds. For other LLMs with smaller context window, they run quicker. 

\vspace{-0.5em}
\paragraph{Expert error analysis} A senior board-certified Emergency Department physician was consulted to provide more in-depth analysis of our generated outputs. This assessment was \textit{not} intended to yield quantitative results but rather to gain insights into how physicians perceive the generated summaries beyond automated metrics and to identify areas for improvement. We focus specifically on discharge summarization and sample 10 pairs of RAG and Direct Generation output from Qwen, as it reached relatively high performance on automated metrics. The physician evaluated the output based on six criteria that have previously been deployed for clinical text evaluation: \textsc{overall accuracy} (factual correctness), \textsc{hallucination},\textsc{omission}, \textsc{readability}, \textsc{clinical relevance} and \textsc{specificity}~\cite{singhal2023large,xu-etal-2024-overview,ben-abacha-etal-2023-investigation,aljamaan2024reference,croxford2024development,williams2024evaluating}. The results of this analysis are given in Figure \ref{fig:human-eval}, and the annotation guidelines can be found in Table~\ref{tab:human_eval}.


Overall, RAG performs slightly better, with fewer hallucinations and more clinically relevant summaries. However, both methods still struggle—often prioritizing less relevant diagnoses, oversimplifying summaries, and failing to discard outdated or disproven diagnoses. These issues highlight ongoing challenges with temporal reasoning.



%% file: appendix.tex
\subsection{More Preprocessing Details}
\label{sec:more_preprocessing}
In addition to the preprocessing steps outlined in section 3.3, we attempt, as far as possible, to only include laboratory measurements and vital sign values that fall outside the normal range. These values are more likely to indicate a problem the patient is facing and are thus more salient. For lab measurements, we only keep those values flagged as 'abnormal'. For chart or vital sign values, there exists a warning flag, which is either 0 or 1. However, some values have an NaN value instead. We exclude values with the flag set to 0, but keep those with 1 or NaN, since it is unknown whether the value corresponding to the NaN type is abnormal or not without a deeper medical analysis.

\subsection{EHRShot Statistics}
\label{sec:ehrshot_stat}
\begin{figure}[ht!]
    \centering
    \includegraphics[width=\columnwidth]{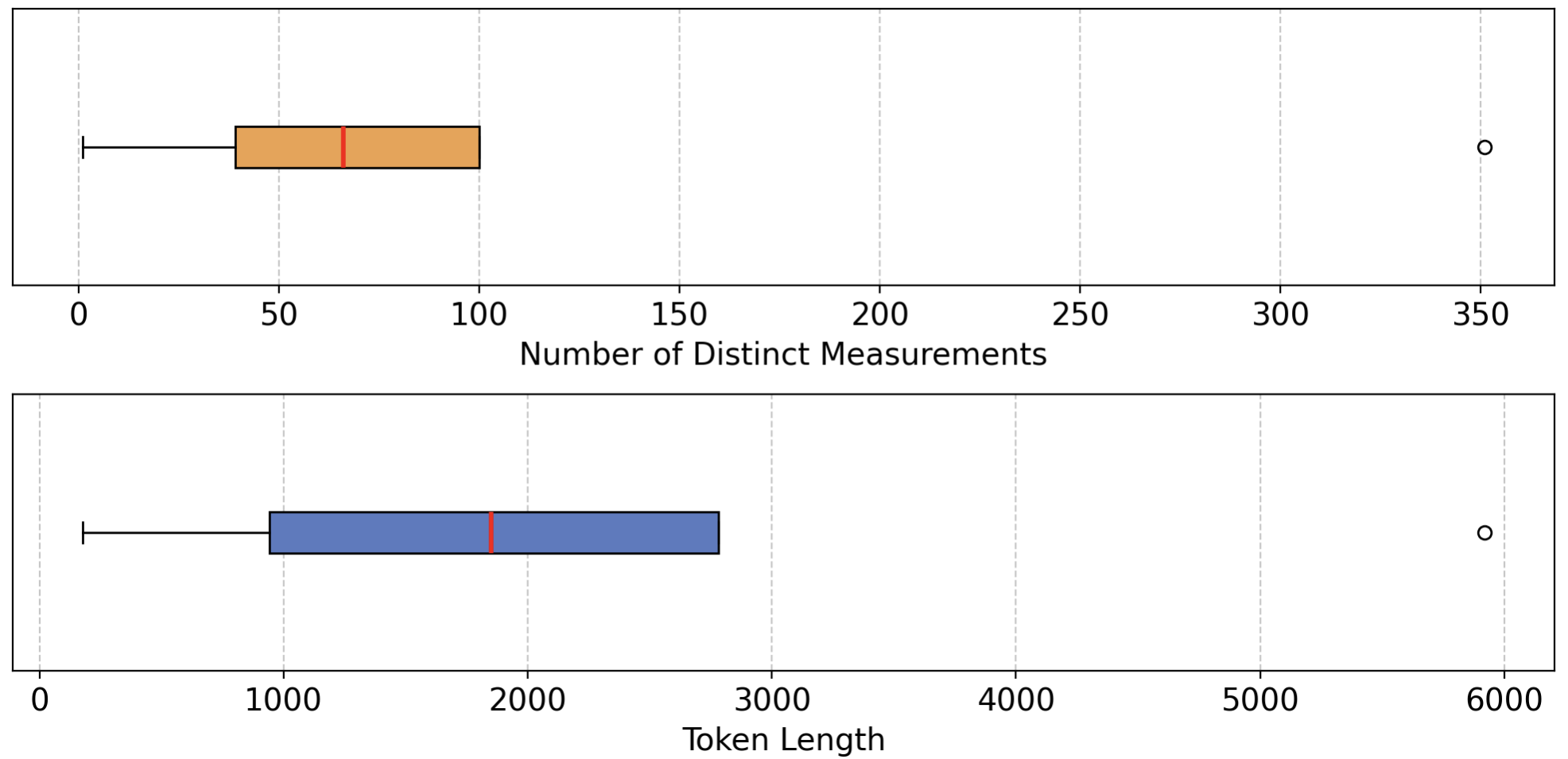}
    \caption{\small Additional statistics on the EHRShot dataset. Number of distinct measurements given above, with the token length distribution below.}
    \label{fig:ehrshot-stat}
\end{figure}

Figure~\ref{fig:ehrshot-stat} reports detailed statistics of EHRShot cohort, regarding the input token length and number of distinct clinical measurements. The prompt token lengths exhibit a wide range, with a mean of 1,989 tokens and a standard deviation of 1,270 tokens. The distribution is right-skewed, with a minimum of 178 tokens and a maximum of 5,919 tokens, while the median (50th percentile) is 1,851 tokens.

For distinct measurement types per patient, the data is also highly variable, with a mean of 74 and a standard deviation of 48. The number of distinct measurements ranges from 1 to 351, with a median of 66 and an interquartile range from 39 (25th percentile) to 100 (75th percentile).

Both distributions highlight significant variation in input complexity, reinforcing the need for models to handle long-context dependencies and multi-modal data effectively.

\begin{table}[]
\small
\centering
\begin{tabular}{@{}lllll@{}}
\toprule
\textbf{Model}  & \textbf{Sec.} & \textbf{CUI}               & \textbf{ROUGE-L}            & \textbf{SapBERT}            \\ \midrule
\multirow{3}{*}{Llama2}  
    & Dx  & 3.87 {\scriptsize ± 5.23}   & 4.01 {\scriptsize ± 3.17}   & 52.21 {\scriptsize ± 11.59}   \\
    & HC  & 7.07 {\scriptsize ± 4.89}   & 12.26 {\scriptsize ± 3.50}  & 61.99 {\scriptsize ± 6.12}    \\
    & DI  & \textbf{10.28} {\scriptsize ± 7.67}  
           & \textbf{13.02} {\scriptsize ± 5.18}  
           & \textbf{63.04} {\scriptsize ± 11.68}   \\ \midrule
\multirow{3}{*}{Mistral}  
    & Dx  & 5.46 {\scriptsize ± 10.07}  & 3.42 {\scriptsize ± 4.34}   & 50.07 {\scriptsize ± 10.78}   \\
    & HC  & 8.64 {\scriptsize ± 5.98}   & \textbf{12.28} {\scriptsize ± 3.13}  
           & \textbf{62.51} {\scriptsize ± 5.95}    \\
    & DI  & \textbf{9.09} {\scriptsize ± 6.90}  
           & 12.07 {\scriptsize ± 3.90}  
           & 60.05 {\scriptsize ± 9.09}   \\ 
\bottomrule
\end{tabular}
\vspace{-0.1in}
\caption{\small Discharge Summarization: Llama2 and Mistral Performance Across Sections: Diagnosis (Dx), Hospital Course (HC), and Discharge Instructions (DI).}
\label{tab:ds_sections}
\end{table}

\subsection{More results on direct generation, RAG and CoT performance comparison }

\begin{table}[h]
\centering
\resizebox{\columnwidth}{!}{%
\begin{tabular}{@{}lllll@{}}
\toprule
\textbf{Model}            & \textbf{Metric} & \textbf{No Prior} & \textbf{Single-Day} & \textbf{Multi-Day} \\ \midrule
\multicolumn{5}{c}{\textit{\textbf{Direct Generation}}}                                                    \\ \midrule
\multirow{4}{*}{Mistral}  & CUI             & 21.60 {\scriptsize± 6.47}      & 34.06 {\scriptsize ± 12.15}       & 30.55 {\scriptsize ± 13.48}      \\
                          & ROUGE           & 17.70 {\scriptsize ± 3.79}      & 26.11 {\scriptsize ± 11.68}       & 23.32 {\scriptsize ± 13.05}      \\
                          & SapBERT         & 68.16 {\scriptsize ± 6.91}      & 71.63 {\scriptsize ± 10.86}       & 72.41 {\scriptsize ± 9.47}       \\
                          & Average         & 35.82 {\scriptsize± 5.72}           & 43.93 {\scriptsize± 11.56}             & 42.09 {\scriptsize± 12}            \\ \midrule
\multirow{4}{*}{Llama3}   & CUI             & 21.83 {\scriptsize ± 6.51}      & 45.35 {\scriptsize ± 10.45}       & 43.55 {\scriptsize ± 12.00}      \\
                          & ROUGE           & 16.60 {\scriptsize ± 3.39}      & 33.16 {\scriptsize ± 10.68}       & 32.42 {\scriptsize ± 12.14}      \\
                          & SapBERT         & 72.01 {\scriptsize ± 6.04}      & 80.10 {\scriptsize ± 8.93}        & 80.42 {\scriptsize ± 8.81}       \\
                          & Average         & 36.81 {\scriptsize ± 4.80}           & \textbf{52.87} {\scriptsize ± 10.02}     & \textbf{52.13}  {\scriptsize ± 10.98}            \\ \midrule
\multirow{4}{*}{Qwen}     & CUI             & 21.84 {\scriptsize ± 5.68}     & 33.8 {\scriptsize ± 6.63}         & 33.96 {\scriptsize ± 6.56}       \\
                          & ROUGE           & 15.15 {\scriptsize ± 3.53}      & 20.74 {\scriptsize ± 4.19}        & 21.6 {\scriptsize ± 5.23}        \\
                          & SapBERT         & 72.17 {\scriptsize ± 4.51}      & 78.01 {\scriptsize ± 4.95}        & 78.41 {\scriptsize ± 4.81}       \\
                          & Average         & 36.38 {\scriptsize ± 4.57}            & 44.18 {\scriptsize ± 5.26}              & 44.65 {\scriptsize ± 5.53}             \\ \midrule
\multirow{4}{*}{DeepSeek} & CUI             & 22.58 {\scriptsize ± 6.63}      & 34.45 {\scriptsize ± 9.26}        & 34.15 {\scriptsize ± 10.00}      \\
                          & ROUGE           & 16.16 {\scriptsize ± 3.15}      & 21.84 {\scriptsize ± 5.65}        & 21.07 {\scriptsize± 5.42}       \\
                          & SapBERT         & 74.97 {\scriptsize ± 3.66}      & 78.29 {\scriptsize ± 4.61}        & 78.35 {\scriptsize ± 5.12}       \\
                          & Average         & \textbf{37.90} {\scriptsize ± 4.48}             & 44.86 {\scriptsize ± 6.51}              & 44.52 {\scriptsize ± 9.51}             \\ \midrule
\multicolumn{5}{c}{\textit{\textbf{RAG}}}                                                                  \\ \midrule
\multirow{4}{*}{Mistral}  & CUI             & 22.65 {\scriptsize± 7.94}      & 35.01 {\scriptsize± 12.40}       & 34.17 {\scriptsize± 10.57}      \\
                          & ROUGE           & 17.15 {\scriptsize± 5.60}      & 29.50 {\scriptsize± 12.46}       & 27.40 {\scriptsize± 10.08}      \\
                          & SapBERT         & 68.57 {\scriptsize± 6.16}      & 74.23 {\scriptsize± 7.79 }       & 73.55 {\scriptsize± 8.46}      \\
                          & Average         & 36.12 {\scriptsize ± 6.57}            & 46.24 {\scriptsize ± 10.88}              & 45.04 {\scriptsize ± 9.70}             \\ \midrule
\multirow{4}{*}{Llama3}   & CUI             & 20.33 {\scriptsize± 5.59}      & 32.02 {\scriptsize± 5.93}        & 31.60 {\scriptsize± 8.38}       \\
                          & ROUGE           & 15.48{\scriptsize ± 3.60 }     & 22.09 {\scriptsize± 5.51}        & 21.42 {\scriptsize± 6.81}       \\
                          & SapBERT         & 73.78 {\scriptsize± 4.50}     & 78.80 {\scriptsize± 4.96}        & 78.96 {\scriptsize± 5.42 }      \\
                          & Average         & 36.53  {\scriptsize ± 4.56}           & 44.3 {\scriptsize ± 5.47}               & 43.99 {\scriptsize ± 6.87}             \\ \midrule
\multirow{4}{*}{Qwen}     & CUI             & 22.07{\scriptsize ± 6.23 }     & 37.3 {\scriptsize± 8.83 }        & 37.21{\scriptsize ± 8.15 }     \\
                          & ROUGE           & 15.6 {\scriptsize± 3.6}        & 25.42 {\scriptsize± 8.08}        & 25.36 {\scriptsize± 7.5}        \\
                          & SapBERT         & 72.16 {\scriptsize± 4.63}      & 79.27 {\scriptsize± 5.37}        & 80.28 {\scriptsize± 4.68}       \\
                          & Average         & 37.46 {\scriptsize ± 4.82}            & 47.33 {\scriptsize ± 7.43}              & 47.45 {\scriptsize ± 6.78}              \\ \bottomrule
\end{tabular}%
}

\vspace{-.1in}
\caption{\small Performance comparison on A\&P generation, aggregated over the patient's entire hospital stay. For readers interested in a more detailed breakdown, we refer them to the Fig~\ref{fig:per_day_cui}, where we provide per-day performance results for a subset of patients with a length of stay of at least 5 days. }
\label{tab:model_performance}
\end{table}

\begin{table}[t]
\small
\centering
\resizebox{\columnwidth}{!}{
\begin{tabular}{@{}llccc@{}}
\toprule
\textbf{Section} & \textbf{Method} & \textbf{CUI} & \textbf{ROUGE-L} & \textbf{BERTScore} \\
\midrule
\multirow{2}{*}{Dx} 
    & Direct Gen & 5.46 {\scriptsize ± 10.07} & \textbf{3.42} {\scriptsize ± 4.34} & \textbf{50.07} {\scriptsize ± 10.78} \\
    & CoT Prompt & 2.80 {\scriptsize ± 5.04}  & 2.95 {\scriptsize ± 3.48} & 48.46 {\scriptsize ± 11.00} \\
\midrule
\multirow{2}{*}{HC} 
    & Direct Gen & \textbf{8.64} {\scriptsize ± 5.98}  & \textbf{12.28} {\scriptsize ± 3.13} & \textbf{62.51} {\scriptsize ± 5.95} \\
    & CoT Prompt & 7.78 {\scriptsize ± 5.89}  & 9.98 {\scriptsize ± 3.84}  & 59.60 {\scriptsize ± 7.32} \\
\midrule
\multirow{2}{*}{DI} 
    & Direct Gen & 9.09 {\scriptsize ± 6.90}  & \textbf{12.07} {\scriptsize ± 3.90} & 60.05 {\scriptsize ± 9.09} \\
    & CoT Prompt & \textbf{9.42} {\scriptsize ± 8.19}  & 10.83 {\scriptsize ± 4.58} & \textbf{60.52} {\scriptsize ± 7.86} \\
\bottomrule
\end{tabular}
}
\vspace{-0.1in}
\caption{\small Mistral performance on the three sections of discharge summarization (Dx: Diagnosis, HC: Hospital course, DI: Discharge Instruction), comparing direct generation approach with the event extraction CoT. Including CUI f-score.}
\label{tab:mistral_direct_vs_cot-cui}
\end{table}

\subsection{Hyperparameter searching results}
\label{sec:hyperparam_tuning}
\begin{table}[h]
\small
\centering
\resizebox{\columnwidth}{!}{%
\begin{tabular}{lccc}
\toprule
\textbf{Experiment} & \textbf{Top-K} & \textbf{Chunk Size} & \textbf{Chunk Overlap} \\
\midrule
\textbf{Exp 1} & 10  & 500  & 100  \\
\textbf{Exp 2} & 20  & 750  & 100  \\
\textbf{Exp 3} & 50  & 500  & 50   \\
\textbf{Exp 4} & 20  & 250  & 100  \\
\textbf{Exp 5} & 50  & 750  & 200  \\
\bottomrule
\end{tabular}%
}
\caption{Hyperparameter configurations for each experiment.}
\label{tab:experiment_settings}
\end{table}

\begin{table}[h]
\small
\centering
\resizebox{\columnwidth}{!}{%

\begin{tabular}{llccccc}
\toprule
\textbf{Model} & \textbf{Metric} & \textbf{Exp 1} & \textbf{Exp 2} & \textbf{Exp 3} & \textbf{Exp 4} & \textbf{Exp 5} \\
\midrule
\multirow{3}{*}{\textbf{Mistral}}
 & ROUGE-L & 6.14 & 6.18 & 5.93 & 6.10 & 6.37 \\
 & SapBERT & 8.41 & 8.12 & 8.37 & 8.35 & 8.03 \\
 & CUI     & 51.03 & 50.82 & 51.12 & 52.56 & 51.34 \\
\midrule
\multirow{3}{*}{\textbf{Llama3}}
 & ROUGE-L & 7.37 & 7.43 & 6.95 & 7.07 & 6.90 \\
 & SapBERT & 9.23 & 9.78 & 9.56 & 9.19 & 9.32 \\
 & CUI     & 55.45 & 56.90 & 56.25 & 55.89 & 56.67 \\
\midrule
\multirow{3}{*}{\textbf{Qwen}}
 & ROUGE-L & 6.41 & 6.97 & 6.95 & 6.60 & 7.11 \\
 & SapBERT & 9.72 & 9.62 & 9.68 & 9.66 & 10.09 \\
 & CUI     & 55.76 & 56.34 & 55.42 & 55.90 & 56.21 \\
\midrule
\multirow{3}{*}{\textbf{Llama2}}
 & ROUGE-L & 7.56 & 7.13 & 7.59 & 7.42 & 6.64 \\
 & SapBERT & 10.22 & 10.21 & 9.96 & 10.16 & 9.41 \\
 & CUI     & 55.67 & 54.70 & 54.37 & 55.80 & 53.50 \\
\bottomrule
\end{tabular}%
}
\caption{Hyperparameter Tuning for Discharge Summaries Results (averaged across sections)}
\label{tab:hyper-ds}
\end{table}

\begin{table}[ht]
\small
\centering
\resizebox{\columnwidth}{!}{%
\begin{tabular}{llccccc}
\toprule
\textbf{Model} & \textbf{Metric} & \textbf{Exp 1} & \textbf{Exp 2} & \textbf{Exp 3} & \textbf{Exp 4} & \textbf{Exp 5} \\
\midrule
\multirow{3}{*}{\textbf{Mistral}} 
 & ROUGE-L & 21.07 & 17.72 & 15.74 & 20.30 & 15.52 \\
 & SapBERT & 71.42 & 70.03 & 68.81 & 71.45 & 69.95 \\
 & CUI     & 28.05 & 21.94 & 20.63 & 27.92 & 19.97 \\
\midrule
\multirow{3}{*}{\textbf{Llama3}}
 & ROUGE-L & 16.73 & 17.54 & 16.74 & 17.11 & 16.58 \\
 & SapBERT & 71.46 & 71.66 & 71.02 & 71.11 & 71.60 \\
 & CUI     & 22.52 & 23.88 & 23.40 & 23.71 & 22.50 \\
\midrule
\multirow{3}{*}{\textbf{Qwen}}
 & ROUGE-L & 19.13 & 21.13 & 20.39 & 18.59 & 20.62 \\
 & SapBERT & 70.29 & 69.34 & 70.98 & 70.22 & 70.03 \\
 & CUI     & 27.28 & 30.17 & 28.87 & 28.04 & 29.87 \\
\bottomrule
\end{tabular}%
}
\caption{Hyperparameter Tuning for A\&P Generation}
\label{tab:hyper-pn}
\end{table}

In this section, we report our hyperparameter tuning on the RAG setup for all LLMs. Table~\ref{tab:experiment_settings} covers the hyperparameter selection. All tuning are done on Discharge Summarization and A\&P generation tasks using a small held-out set (n=5). The best set of parameters are used for official experiments reported in the main text.

\subsection{Prompt and query optimization}
\label{sec:prompt_opt}
\subsection{Event extraction COT prompt for discharge summarization }

\begin{figure}[t]
\centering
\small
\begin{tcolorbox}[title=\textbf{Chain-of-Thought Prompt for 48h Discharge Summary Event Extraction}, colback=gray!5, colframe=gray!40!black, width=\columnwidth]
\textbf{DISCHARGE EVENT EXTRACTION TASK}
Analyze the following data from the final 48 hours of the hospital stay and identify key clinical events that are most relevant for summarizing the course of treatment and informing discharge planning.
\{chronology\_text\}
Only include events that reflect:
\begin{enumerate}[nosep, leftmargin=*]
    \item Significant changes in symptoms or status (e.g., improvements, worsening, new findings).
    \item Clinically important test results (especially abnormal values that lead to certain treatments).
    \item Major treatments or interventions (e.g., medication changes, procedures, escalation/de-escalation).
    \item Care team decisions that indicate readiness for discharge or change in care goals.
    \item Events linked to the final diagnosis or that inform follow-up care.
\end{enumerate}
\textbf{Response Format:}
\#\#\# Day X Key Events \#\#\#
- [Time]: [Description] (Reasoning)
\end{tcolorbox}
\vspace{-.1in}
\begin{tcolorbox}[title=\textbf{Example Model Output}, colback=blue!2, colframe=blue!50!black, width=\columnwidth]
\#\#\# Day 1 Key Events \#\#\#
- \textbf{**2159-03-12 08:00**}: Stable Vital Signs (BP 120/70 mmHg, HR 88 bpm, RR 18 breaths/min, Temp 98.6°F). This indicates overall stability and readiness for discharge.
\end{tcolorbox}
\caption{\small Chain-of-Thought prompting template and example output used for discharge summary event extraction based on the final 48 hours of hospitalization.}
\label{fig:cot_discharge_prompt}
\end{figure}

\subsection{RAG and direct generation prompt optimization}
We optimized the prompts and queries used for RAG systems on all tasks. Both prompts and queries are similar in terms of its instruction, the only difference is that queries for RAG has the ``Retrieve" component.

\begin{table*}[ht]
\small
\centering
\begin{tabular}{|l|}
\hline
What is the patient’s main diagnosis?                                                                                         \\ \hline
\textbf{The patient’s primary diagnosis is: (Llama2)}                                                                                  \\ \hline
\textbf{Identify the primary reason for the patient’s hospital admission: (Llama3)}                                                           \\ \hline
\textbf{Instruct: Given a search query, retrieve relevant passages that answer the query.}\\ \textbf{Query: patient’s primary diagnosis. (Mistral)} \\ \hline
\textbf{The patient has been diagnosed with: (Qwen)}                                                                                 \\ \hline
\end{tabular}
\caption{Queries used for retrieving the primary diagnosis. Highest performing on models bolded with respective model in parentheses.}
\label{tab:diagnosis-qs}
\end{table*}

\begin{table*}[ht]
\small
\centering
\begin{tabular}{|l|}
\hline
\textbf{Summarize the hospital course for this patient in a concise and accurate way.} \\ \textbf{(Qwen)}                                           \\ \hline
The patient’s hospital course included the following:                                                          \\ \hline
\textbf{Provide a brief hospital course, including key events and treatments.}\\
\textbf{(Mistral, Llama3)}                                          \\ \hline
Instruct: Given a search query, retrieve relevant passages that answer the query. \\
Query: Brief hospital course. \\ \hline
\textbf{What were the key events and outcomes during the patient’s hospital stay?}
\\ \textbf{(Llama2)} \\ \hline
\end{tabular}
\caption{Queries used for retrieving the hospital course. Highest performing on models bolded with respective model in parentheses.}
\label{tab:bhc-qs}
\end{table*}

\begin{table*}[ht]
\small
\centering
\begin{tabular}{|l|}
\hline
\textbf{Given the input EHR data, generate discharge instructions for this patient. (All models)}                                             \\ \hline
What are the discharge instructions for the patient?                                                                    \\ \hline
Write a summary of the discharge plan, including medications, follow-up visits, and \\ patient care instructions. \\ \hline
Instruct: Given a search query, retrieve relevant passages that answer the query. \\
Query: discharge instructions.         \\ \hline
What follow-up care and medications are recommended for the patient after discharge?                          \\ \hline
\end{tabular}
\caption{Queries used for retrieving the discharge instructions. Highest performing on models bolded with respective model in parentheses.}
\label{tab:discharge-qs}
\end{table*}

\begin{table*}[ht]
\small
\centering
\begin{tabular}{|l|}
\hline
\begin{tabular}[c]{@{}l@{}}Given the patient EHR data, write the Assessment section of a clinical progress note. The Assessment should include \\ a brief description of both passive and active diagnoses. Clearly state why the patient is admitted to the hospital \\ and describe the active problem for the day, along with any relevant comorbidities the patient has.\end{tabular} \\ \hline
\begin{tabular}[c]{@{}l@{}}Provide the Assessment section of the patient’s progress note, including active and passive diagnoses, admission reasons, \\ the patient’s active problems for the day, and relevant comorbidities.\end{tabular}                                                                                                                                               \\ \hline
\begin{tabular}[c]{@{}l@{}}What are the patient’s active and passive diagnoses? Why was the patient admitted to the hospital? \\ What are the active medical problems for the day? Include relevant comorbidities.\end{tabular}                                                                                                                                                           \\ \hline
\textbf{\begin{tabular}[c]{@{}l@{}}Retrieve passages that explain the patient's active and passive diagnoses, reasons for admission, active problems \\ for the day, and relevant comorbidities. (Mistral, Qwen)\end{tabular}}                                                                                                                                                            \\ \hline
\textbf{\begin{tabular}[c]{@{}l@{}}Instruct: Generate a concise Assessment section for the patient’s progress note. Include a summary of active and\\  passive diagnoses, admission reasons, the patient’s current active problems, and any comorbidities. (Llama3)\end{tabular}}                                                                                                         \\ \hline
\end{tabular}
\caption{Queries used for retrieving the Assessment section of a progress note. Highest performing on models bolded with respective model in parentheses.}
\label{tab:query-a}
\end{table*}

\begin{table*}[ht]
\small
\centering
\begin{tabular}{|l|}
\hline
\begin{tabular}[c]{@{}l@{}}Given the patient EHR data, write the Plan section of a clinical progress note. The Plan should be organized into multiple \\ subsections, each corresponding to a specific medical problem. Provide a detailed treatment plan for each problem, outlining \\ proposed or ongoing interventions, medications, and care strategies.\end{tabular} \\ \hline
\begin{tabular}[c]{@{}l@{}}Generate the Plan section of the patient’s progress note. Provide detailed treatment plans for specific medical problems, \\ including proposed or ongoing interventions, medications, and care strategies.\end{tabular}                                                                                                                        \\ \hline
\begin{tabular}[c]{@{}l@{}}What are the proposed treatment plans for the patient’s active medical problems? Include details on interventions, \\ medication regimens, and care strategies.\end{tabular}                                                                                                                                                                    \\ \hline
\textbf{\begin{tabular}[c]{@{}l@{}}Retrieve passages that outline treatment strategies for medical problems, including medications, interventions, \\ and care strategies. \\ (Mistral, Qwen)\end{tabular}}                                                                                                                                                                   \\ \hline
\textbf{\begin{tabular}[c]{@{}l@{}}Instruct: Write the Plan section of the progress note. Organize it into subsections for each medical problem. \\ Provide detailed plans for treatments, interventions, medications, and care strategies. (Llama3)\end{tabular}}                                                                                                         \\ \hline
\end{tabular}
\caption{Queries used for retrieving the Plan section of a progress note. Highest performing on models bolded with respective model in parentheses.}
\label{tab:query-p}
\end{table*}

Table~\ref{tab:diagnosis-qs},\ref{tab:bhc-qs},\ref{tab:query-a},\ref{tab:query-p},\ref{tab:discharge-qs} presents all prompts and queries for MIMIC tasks. 


\subsection{Impact of context length on consecutive patient data}
\begin{figure}
    \centering
    \includegraphics[width=\columnwidth]{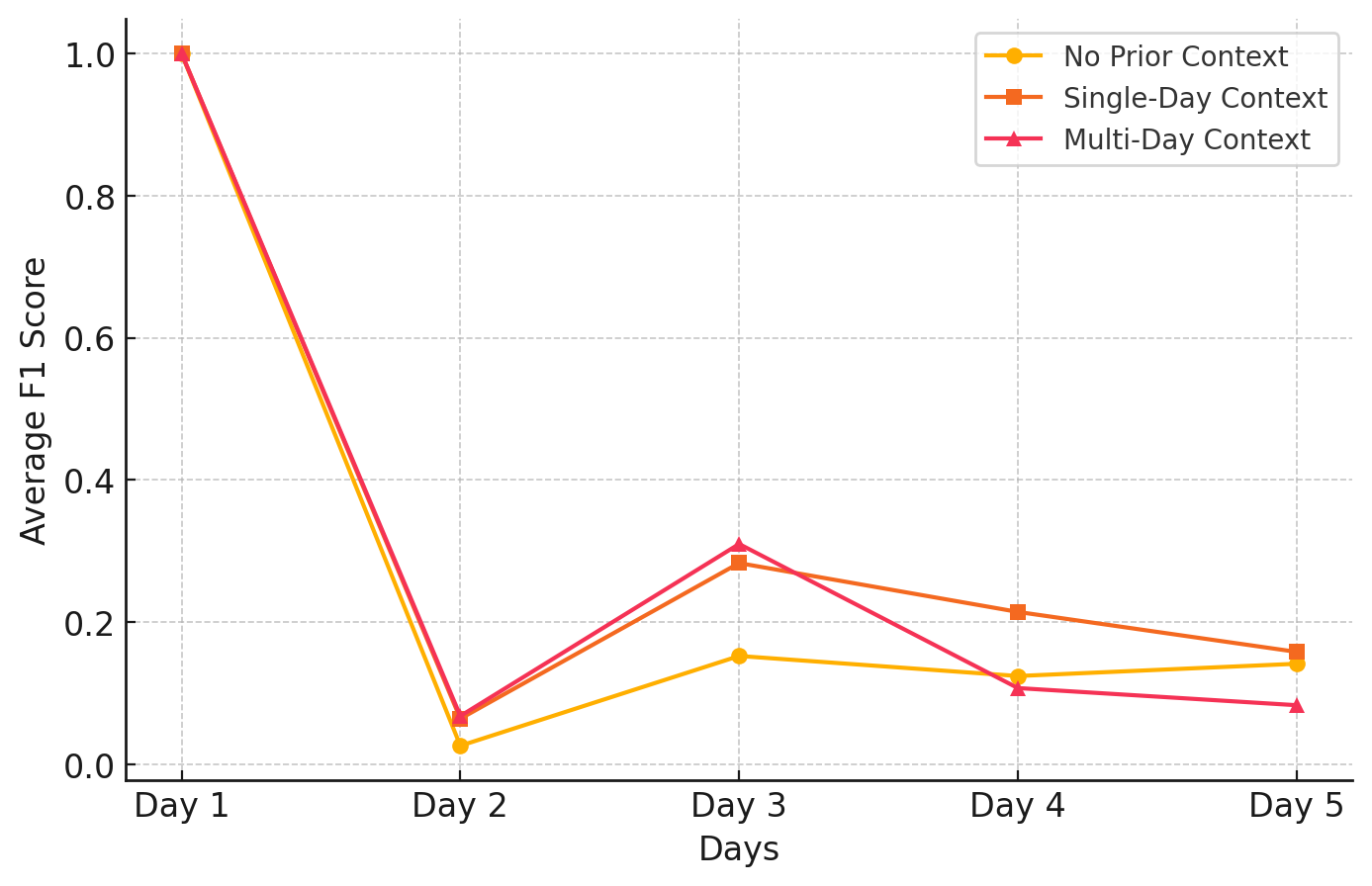}
    \caption{\small Impact of prior-day context on F1 score across consecutive days}
    \label{fig:per_day_cui}
\end{figure}

Figure~\ref{fig:per_day_cui} illustrates the impact of incorporating prior-day context on F1 scores for patient note generation. The x-axis represents consecutive relative days within a patient's hospital stay, where Day 0 corresponds to the first day of admission and serves as the ground truth (i.e., no evaluation is performed on this day). The y-axis shows the average F1 score across a subset of 10 patients with a length of stay of at least 5 days.

Since Day 0 (Day 1 in the figure) is not evaluated, its F1 score is set to 1.0 for all methods to ensure a clear visual comparison with subsequent days. The results demonstrate that incorporating prior-day context improves performance over time. The Single-Day Context and Multi-Day Context methods achieve substantially higher F1 scores than the No Prior Context method, particularly on Day 2 and Day 3, suggesting that leveraging past information helps generate more accurate and coherent patient notes. However, after Day 3, the performance of the Multi-Day Context method begins to decline, indicating that while longer historical context can be beneficial, it may introduce additional noise or redundant information.

Overall, these findings highlight that considering prior context enhances the accuracy of generated patient notes, with Single-Day Context yielding the highest performance in later days, while Multi-Day Context shows initial improvements but exhibits diminishing returns over time.

\begin{table*}[ht]
\small
\centering
\begin{tabular}{llllllll}
\hline
            &  & \multicolumn{3}{c}{\textbf{GT}}                                                                                                                                                         & \multicolumn{3}{c}{\textbf{GEN}}                                                                                                                                                        \\ 
                  \textbf{Model}       &    \textbf{Metric}             & \begin{tabular}[c]{@{}l@{}}\textbf{No Prior}\end{tabular} & \begin{tabular}[c]{@{}l@{}}\textbf{Single-Day}\end{tabular} & \begin{tabular}[c]{@{}l@{}}\textbf{Multi-Day}\end{tabular} & \begin{tabular}[c]{@{}l@{}}\textbf{No Prior}\end{tabular} & \begin{tabular}[c]{@{}l@{}}\textbf{Single-Day} \end{tabular} & \begin{tabular}[c]{@{}l@{}}\textbf{Multi-Day} \end{tabular} \\ \hline
\multirow{4}{*}{Mistral}  & CUI             & 21.60 ± 6.47                                            & 34.06 ± 12.15                                                & 30.55 ± 13.48                                               & 21.60 ± 6.47                                               & 22.30 ± 16.13                                                & 22.82 ± 16.47                                               \\
                          & ROUGE-L         & 17.70 ± 3.79                                               & 26.11 ± 11.68                                                & 23.32 ± 13.05                                               & 17.70 ± 3.79                                               & 17.57 ± 13.47                                                & 17.90 ± 13.59                                               \\
                          & SapBERT         & 68.16 ± 6.91                                               & 71.63 ± 10.86                                                & 72.41 ± 9.47                                                & 68.16 ± 6.91                                               & 70.51 ± 11.60                                                & 71.50 ± 9.66                                                \\
                          & Average         & 35.82                                                      & 43.93                                                        & 42.09                                                       & 35.82                                                      & 36.79                                                        & 37.41                                                       \\ \hline
\multirow{4}{*}{Llama3}   & CUI             & 21.83 ± 6.51                                               & 45.35 ± 10.45                                                & 43.55 ± 12.00                                               & 21.83 ± 6.51                                               & 40.89 ± 13.61                                                & 40.50 ± 13.29                                               \\
                          & ROUGE-L         & 16.60 ± 3.39                                               & 33.16 ± 10.68                                                & 32.42 ± 12.14                                               & 16.90 ± 3.85                                               & 28.86 ± 12.8                                                 & 29.34 ± 13.30                                               \\
                          & SapBERT         & 72.01 ± 6.04                                               & 80.10 ± 8.93                                                 & 80.42 ± 8.81                                                & 72.33 ± 5.48                                               & 79.68 ± 10.07                                                & 80.48 ± 9.56                                                \\
                          & Average         & 36.81                                                      & 52.87                                                        & 52.13                                                       & 37.05                                                      & 49.81                                                        & 50.11                                                       \\ \hline
\multirow{4}{*}{Qwen}     & CUI             & 21.84 ± 5.68                                               & 34.27 ± 7.18                                                 & 34.32 ± 6.65                                                & 21.84 ± 5.68                                               & 30.10 ± 6.70                                                 & 29.55 ± 6.50                                                \\
                          & ROUGE-L         & 15.35 ± 3.10                                               & 77.50 ± 5.37                                                 & 21.85 ± 4.99                                                & 15.18 ± 3.00                                               & 18.21 ± 4.28                                                 & 17.55 ± 4.07                                                \\
                          & SapBERT         & 72.57 ± 4.92                                               & 20.69 ± 4.37                                                 & 77.88 ± 4.86                                                & 72.97 ± 4.83                                               & 78.39 ± 4.11                                                 & 77.61 ± 4.52                                                \\
                          & Average         & 36.87                                                      & 44.15                                                        & 44.68                                                       & 37.12                                                      & 42.23                                                        & 41.57                                                       \\ \hline
\multirow{4}{*}{DeepSeek} & CUI             & 22.58 ± 6.63                                               & 34.45 ± 9.26                                                 & 34.15 ± 10.00                                               & 22.58 ± 6.63                                               & 27.33 ± 7.69                                                 & 27.93 ± 7.46                                                \\
                          & ROUGE-L         & 16.16 ± 3.15                                               & 21.84 ± 5.65                                                 & 21.07 ± 5.42                                                & 16.16 ± 3.15                                               & 17.37 ± 3.86                                                 & 16.88 ± 3.88                                                \\
                          & SapBERT         & 74.97 ± 3.66                                               & 78.29 ± 4.61                                                 & 78.35 ± 5.12                                                & 74.97 ± 3.66                                               & 77.90 ± 4.36                                                 & 77.93 ± 4.79                                                \\
                          & Average         & 37.90                                                      & 44.86                                                        & 44.52                                                       & 37.90                                                      & 40.87                                                        & 40.91                                                       \\ \hline
\end{tabular}
\caption{Comparison of ground-truth (GT, same results we report in main text) and generated (GEN) settings on assessment and plan generation (results of direct generation approach shown)}
\label{tab:gt-gen}
\end{table*}

\begin{figure*}
    \centering
    \includegraphics[width=\textwidth]{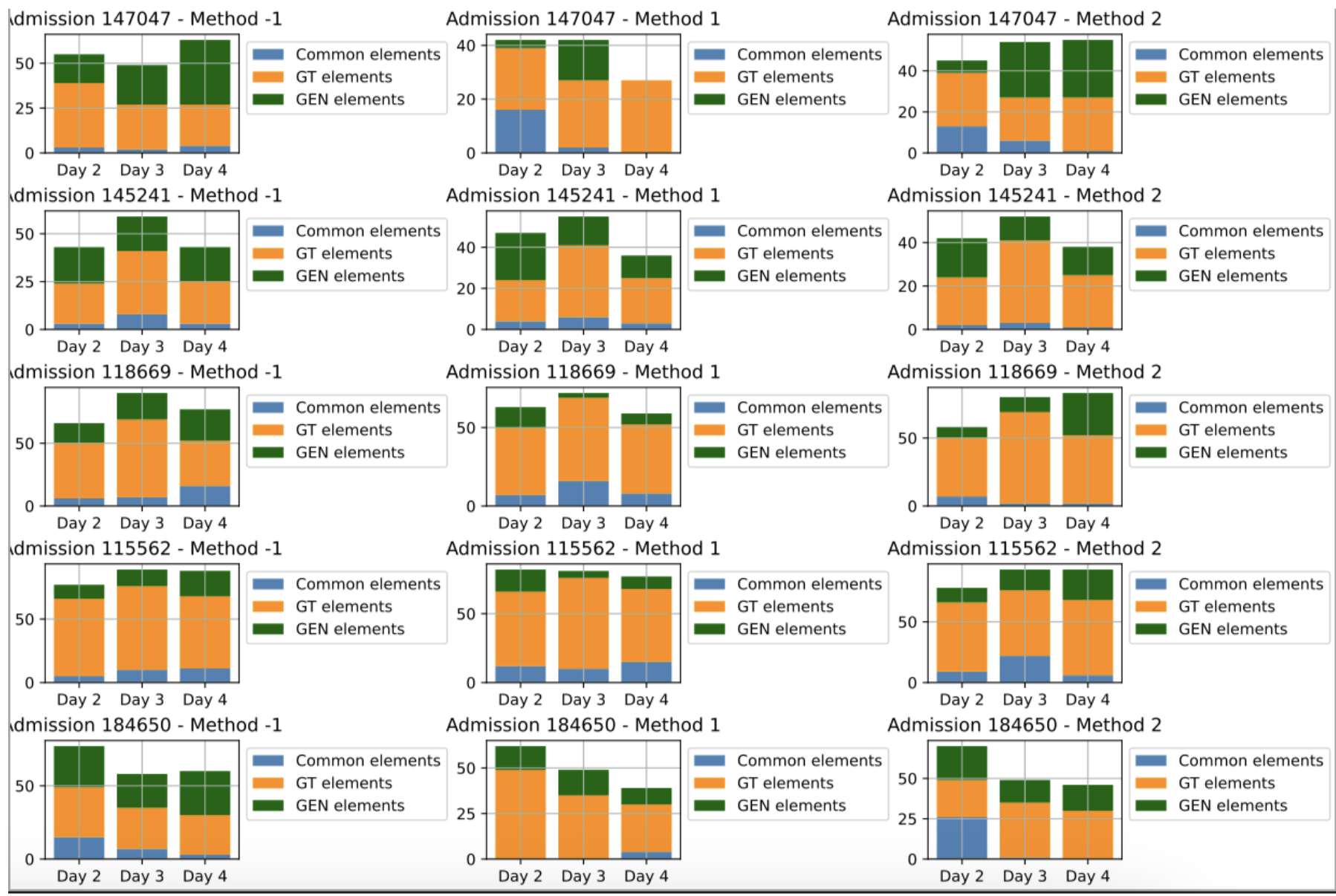}
    \caption{\small Comparison of generated patient notes across different methods of incorporating prior-day context. Each subplot represents a different patient admission, with bars indicating the composition of generated notes across Days 2, 3, and 4. The stacked bars show the proportion of common elements (overlapping between ground truth and generated notes), GT elements (present in the ground truth but missing in generated notes), and GEN elements (newly generated content not found in the ground truth). Method -1 (no prior-day context), Method 1 (single prior-day context), and Method 2 (multi-day context) demonstrate how historical context influences the balance between accurate retention and newly introduced information.}
    \label{fig:pn_as_day_goes}
\end{figure*}

\begin{figure*}[ht]
    \centering
    \includegraphics[width=\textwidth]{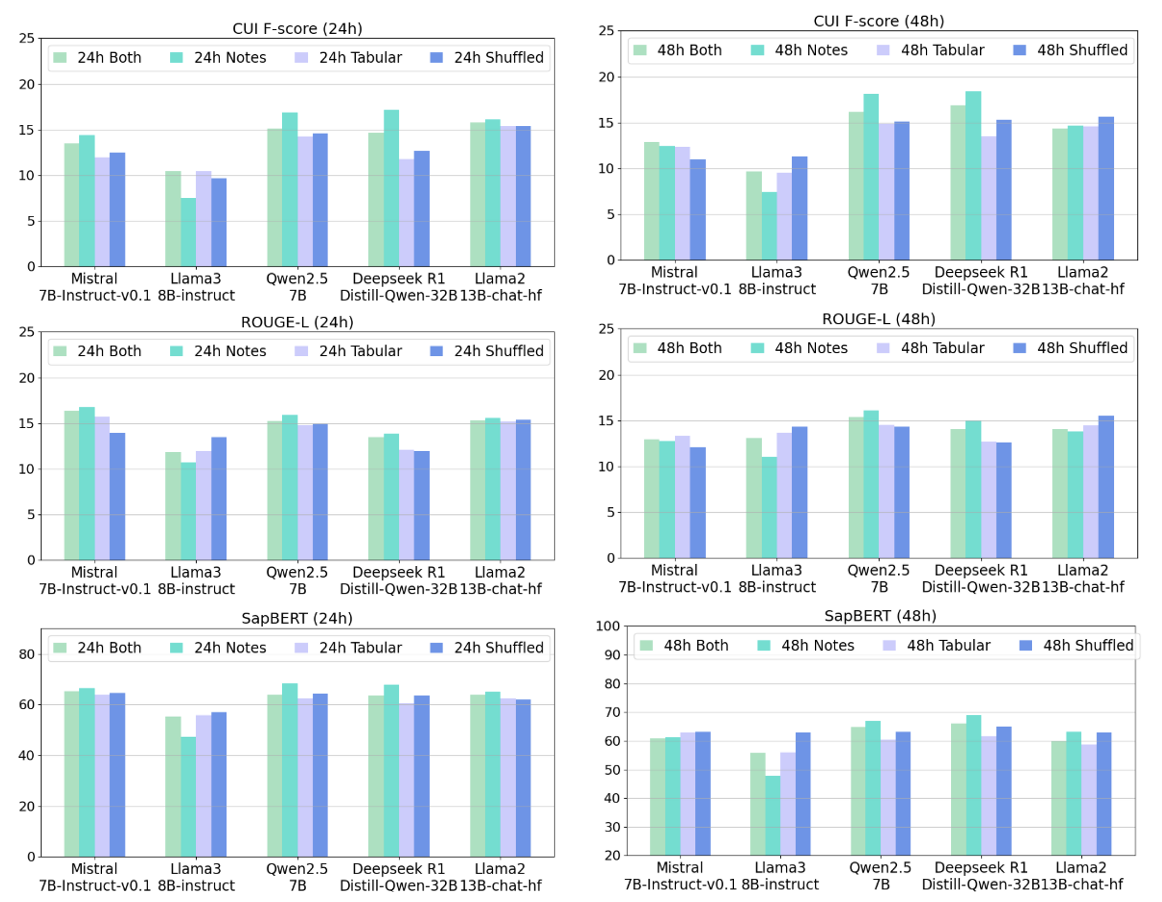}
    \caption{\small Metric breakdown across modalities for the 24 hour time window (on direct generation)}
    \label{fig:24h-48h-metrics}
\end{figure*}

\subsection{More results for using ground-truth progress notes vs. generated progress notes on A\&P Generation}

This section presents further results comparing ground-truth progress notes and model-generated progress notes as input for A\&P Generation. Our analysis evaluates how using prior ground-truth notes versus LLM-generated notes impacts the overall quality, coherence, and accuracy of the generated A\&P sections.

Table~\ref{tab:gt-gen} presents results on all LLMs direct generation. Overall, there are performance decreases when moving from using generated progress notes as context input. The results highlight key differences in information retention and propagation, where models using ground-truth progress notes tend to maintain better clinical consistency, while those using generated notes may accumulate errors over multiple days, leading to drift and hallucination in longitudinal patient summaries. These findings emphasize the need for robust calibration and error correction mechanisms when relying on LLM-generated progress notes for iterative summarization.

Additionally, we include Figure~\ref{fig:pn_as_day_goes} to show the impact of incorporating prior-day context on the composition of generated patient notes. The figure illustrates how different methods—ranging from no prior-day context (Method -1) to single-day (Method 1) and multi-day context (Method 2)—affect the alignment between generated elements (GEN), ground truth elements (GT), and common elements across multiple admissions. As the number of prior days included increases, we observe changes in the proportion of correctly retained ground truth elements and newly generated content, highlighting the role of historical context in improving consistency and completeness in patient note generation. Additionally, the inclusion of prior context reduces the number of newly generated elements, indicating a shift away from generating potentially extraneous or less relevant content.

However, the degree of improvement varies across different admissions, suggesting that some cases benefit more from historical context than others, potentially due to differences in case complexity or the structure of prior notes. Furthermore, the differences between generated and ground truth elements become more pronounced as days progress, highlighting the challenge of maintaining consistency in patient notes over time. Overall, the findings suggest that incorporating multi-day context enhances the accuracy and stability of generated patient notes, reducing hallucinated content while preserving clinically relevant information.

\begin{table*}[ht]
\small 
\centering
\begin{tabularx}{\textwidth}{p{3.5cm}Xc}
\toprule
\multicolumn{3}{l}{\textbf{Physician Evaluation of LLM-Generated Clinical Summaries}}\\ 
\multicolumn{3}{p{\textwidth}}{Thank you for participating in this evaluation. Please assess the LLM-generated text based on the following criteria using the 5-point Likert scale provided for each question.} \\ 
\multicolumn{3}{l}{Scale Definitions (from 1-5):} \\
\multicolumn{3}{l}{1 Strongly Disagree; 2 Disagree; 3 Neutral; 4 Agree; 5 Strongly Agree}
 \\ 
 \midrule 
\textbf{Category} & \textbf{Evaluation Question} & \textbf{Score (1-5)} \\
\midrule
\textbf{Overall Accuracy}~\cite{xu-etal-2024-overview,singhal2023large,ben-abacha-etal-2023-investigation,croxford2024development,johri2025evaluation} & How well does the generated text align with the actual clinical data? \newline
The summary is factually correct and accurately represents the original data. \newline
No major distortions or misinterpretations of key clinical facts. &  \\
\midrule
\textbf{Hallucination (Faithfulness)}~\cite{ben-abacha-etal-2023-investigation,singhal2023large, aljamaan2024reference,johri2025evaluation} & To what extent does the LLM generate information faithfully? \newline
The generated text does not introduce any fabricated, misleading, or incorrect information. \newline
All statements in the summary can be traced back to the original document. & \\
\midrule
\textbf{Omission}~\cite{croxford2024development,ben-abacha-etal-2023-investigation} & Did the model include all important clinical details? \newline
The generated summary includes all clinically important details. \newline
No missing critical pieces of information relevant to patient care. &   \\
\midrule
\textbf{Readability}~\cite{xu-etal-2024-overview} & How easy is the generated text to read and comprehend? \newline
The text is well-structured, clear, and easy to understand. \newline
Uses appropriate medical terminology without excessive complexity. &  \\
\midrule
\textbf{Clinical Relevance} (Usefulness)~\cite{singhal2023large,johri2025evaluation} & How useful is the content for clinical decision-making? \newline
The information provided is highly relevant to the clinical task. \newline
Avoids unnecessary or unrelated details. & \\
\midrule
\textbf{Specificity (Level of Detail)}~\cite{williams2024evaluating} & Does the summary maintain an appropriate level of detail? \newline
The text balances between a high-level summary and necessary details. \newline
Avoids being overly vague or excessively detailed. &   \\
\bottomrule
\end{tabularx} 
\caption{\small Physician evaluation survey}
\label{tab:human_eval}
\end{table*}


\subsection{Break-down results behind Figure~\ref{fig:average_f1}} 

Recall that Figure~\ref{fig:average_f1} aggregates all metrics for LLM direct generation on discharge summarization. To provide comprehensive results analysis, we include Figure~\ref{fig:24h-48h-metrics} for 24h window and 48h window. The trends of performance changing across different modalities are consistent with the Figure~\ref{fig:average_f1}. 

\subsection{Details about human evaluation}
\label{sec:human_eval}
Last but not least, we present the full survey questions aggregated from existing work in Table~\ref{tab:human_eval}. We would like to emphasize that this survey has \textit{not} been validated, rather, it selects the criteria after consulting with a physician regarding what they value in LLM generated text and findings from prior work, as cited in the table. These questions allow us to evaluate the accuracy, faithfulness, readability, and clinical relevance of LLM-generated text, providing a structured framework for assessing their strengths and limitations in a clinical setting. 